\documentclass[10pt,twocolumn,letterpaper]{article}

\usepackage[pagenumbers]{iccv} 

\renewcommand\footnotemark{}

\usepackage{times}
\usepackage{epsfig}
\usepackage{graphicx}
\usepackage{amsmath}
\usepackage{amssymb}
\usepackage{booktabs}
\usepackage{array}
\usepackage{xcolor}
\usepackage{multirow}
\usepackage{tabularx}
\usepackage{siunitx}
\usepackage{cprotect}
\usepackage{svg}
\usepackage{stfloats}
\usepackage{afterpage}

\makeatletter

\newcommand*{\@rowstyle}{}

\newcommand*{\rowstyle}[1]{
  \gdef\@rowstyle{#1}%
  \@rowstyle\ignorespaces%
}

\newcolumntype{=}{
  >{\gdef\@rowstyle{}}%
}

\newcolumntype{+}{
  >{\@rowstyle}%
}

\makeatother

\definecolor{iccvblue}{rgb}{0.21,0.49,0.74}
\usepackage[pagebackref,breaklinks,colorlinks,allcolors=iccvblue]{hyperref}

\def\paperID{11351} 
\def\confName{ICCV}
\def\confYear{2025}

\title{SuperEvent:~\hspace{-0.75pt}Cross-Modal~\hspace{-0.75pt}Learning~\hspace{-0.75pt}of~\hspace{-0.75pt}Event-based~\hspace{-0.75pt}Keypoint~\hspace{-0.75pt}Detection~\hspace{-0.75pt}for~\hspace{-0.75pt}SLAM}

\author{Yannick Burkhardt$^{1,2,3}$
\quad
Simon Schaefer$^{1,2,3}$
\qquad
Stefan Leutenegger$^{1,2,3}$%
\thanks{$^{1}$Technical University of Munich,}
\thanks{~~\{\href{mailto:yannick.burkhardt@tum.de}{\nolinkurl{yannick.burkhardt}}, \href{mailto:simon.k.schaefer@tum.de}{\nolinkurl{simon.k.schaefer}} \}\href{tum.de}{\nolinkurl{@tum.de}} }
\thanks{$^{2}$ETH Zürich, \href{mailto:lestefan@ethz.ch}{\nolinkurl{lestefan@ethz.ch}}}
\thanks{$^{3}$Munich Center for Machine Learning (MCML)}}

\begin{document}
\maketitle
\begin{abstract}
    Event-based keypoint detection and matching holds significant potential, enabling the integration of event sensors into highly optimized Visual SLAM systems developed for frame cameras over decades of research. Unfortunately, existing approaches struggle with the motion-dependent appearance of keypoints and the complex noise prevalent in event streams, resulting in severely limited feature matching capabilities and poor performance on downstream tasks. To mitigate this problem, we propose SuperEvent, a data-driven approach to predict stable keypoints with expressive descriptors. Due to the absence of event datasets with ground truth keypoint labels, we leverage existing frame-based keypoint detectors on readily available event-aligned and synchronized gray-scale frames for self-supervision: we generate temporally sparse keypoint pseudo-labels considering that events are a product of both scene appearance and camera motion. Combined with our novel, information-rich event representation, we enable SuperEvent to effectively learn robust keypoint detection and description in event streams. Finally, we demonstrate the usefulness of SuperEvent by its integration into a modern sparse keypoint and descriptor-based SLAM framework originally developed for traditional cameras, surpassing the state-of-the-art in event-based SLAM by a wide margin. Source code is available at \href{https://ethz-mrl.github.io/SuperEvent/}{\nolinkurl{ethz-mrl.github.io/SuperEvent}}.
\end{abstract}
\section{Introduction}
\label{sec:intro}

\begin{figure}
  \includegraphics[width=\columnwidth]{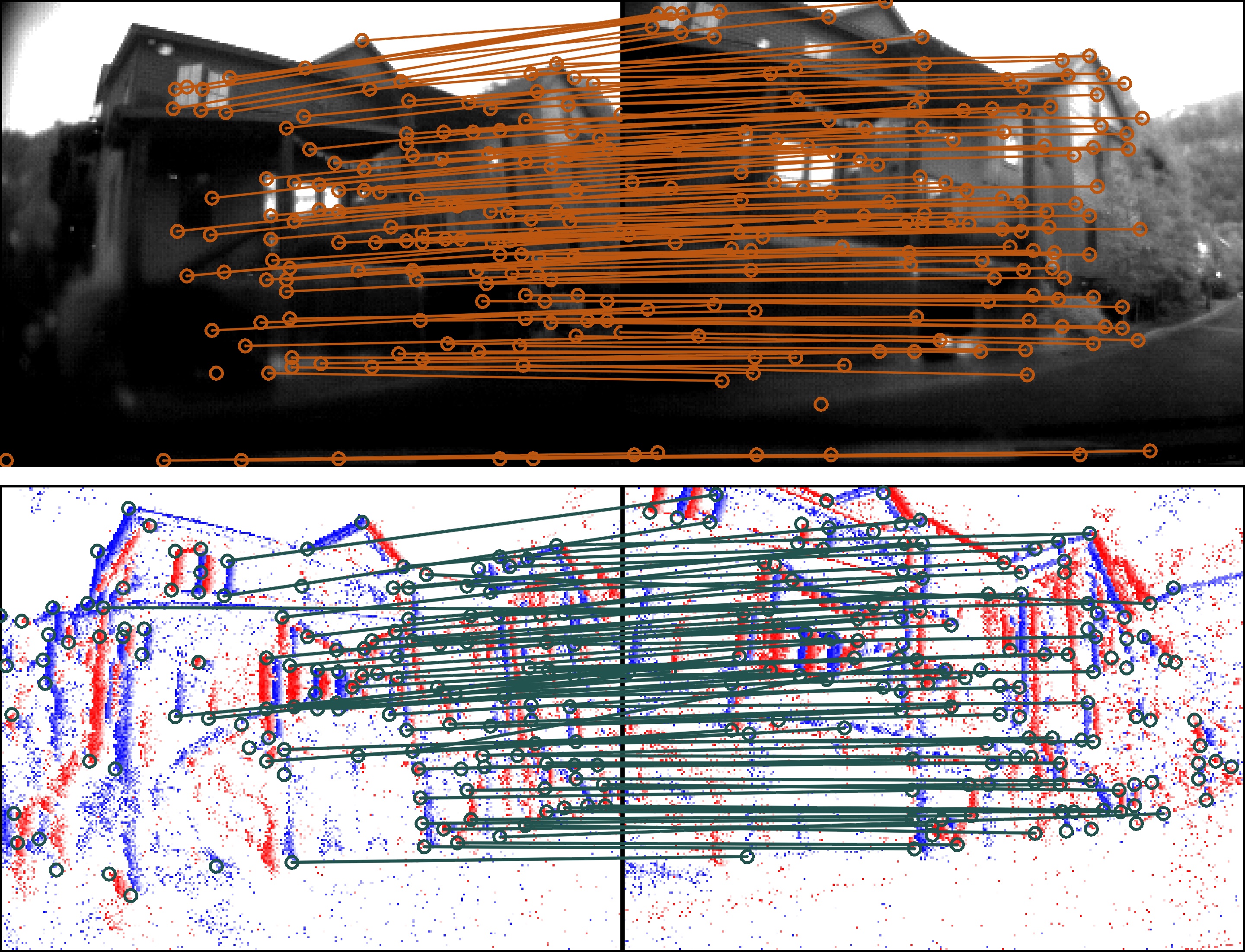}
  \caption{Detections on the sequence \textit{rec1499023756} of the DDD20~\cite{Hu20} dataset (not used for training). Top: Pseudo-labels from SuperPoint~\cite{Det18} and SuperGlue~\cite{Sar20} on the gray-scale frames. Bottom: Matched keypoints from SuperEvent in the event stream at the corresponding time stamps.}
  \label{teaser}
  \vspace{-2pt}
\end{figure}

\begin{figure*}[t]
  \includegraphics[width=\textwidth]{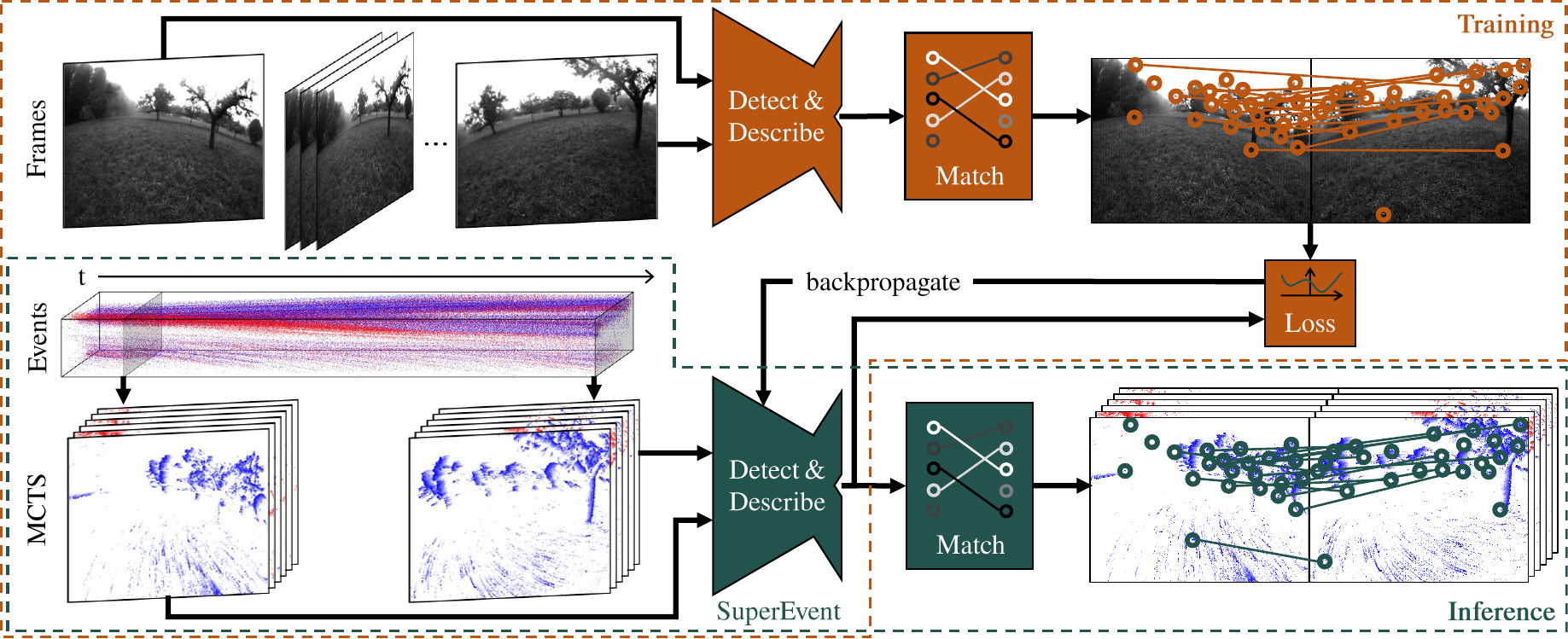}
  \caption{Data processing pipeline of SuperEvent: For training, pseudo-labels are generated using a frame-based detector and matcher on the gray-scale frame pairs. We generate spatially synchronized MCTS at the same timestamps and feed them to SuperEvent. The network predictions are compared to the pseudo-labels and the network weights are optimized using backpropagation. During inference, the network predictions can be used to detect keypoints and match their descriptors on the event stream only.}
  \label{pipeline}
  \vspace{-6pt}
\end{figure*}

Event cameras offer exciting advantages over their frame-based counterparts, such as an increased temporal resolution, little motion blur, and a high dynamic range. These properties promise to enhance robustness for robot estimation and perception tasks under fast motion and uncontrollable lighting conditions. However, processing the sparse and asynchronous output of event cameras requires fundamentally different algorithms than traditional frames. The decades-long research edge in frame-based vision resulted in most hardware and software being highly optimized for frame processing. A common approach for tasks like Structure-from-Motion, Visual Odometry (VO), Simultaneous Localization and Mapping (SLAM), or Place Recognition is to detect and match keypoints across frames to gain a geometric scene understanding~\cite{Leu22, Cam21}. Consequently, a robust keypoint detector and descriptor working on the event stream would unlock the potential of integrating event cameras into extensive and highly-developed algorithms developed initially for frame-based vision.\par
While several authors already proposed event-based approaches to detect keypoints~\cite{Man19, Chi21, Chi22, Hua23, Iku24, Gao24}, their matching abilities still show limitations. Most employ nearest-neighbor matching in pixel space, which restricts their usability to visually simple scenes. In an attempt to overcome this issue, more recent approaches learn descriptors in a data-driven fashion. To generate ground truth labels -- inspired by frame-based methods~\cite{Det18} -- they warp their event representations with a random homography and penalize non-consistent model predictions. This approach forces the models to learn descriptors to match visually similar keypoints. However, it fundamentally ignores the true nature of the event generation process that conflates camera motion with scene appearance and geometry. In essence, matches are generated only from one original set of events per keypoint, thereby producing only limited and potentially also unrealistic training data. We hypothesize that this simplification ultimately results in unstable matching.\par
In contrast, we propose training our model to predict descriptors for the keypoint correspondences at \emph{different} points in time -- therefore training with \emph{different} and actually \emph{recorded} rather than warped events. Since generating keypoint and match labels manually is intractable, we design a scalable method to exploit the great capabilities of frame-based vision models: generating pseudo-labels in aligned frames and then using them to train our model's predictions. An example is shown in Figure~\ref{teaser}.\par
To provide the model with rich input data, we propose the novel event representation Multi-Channel Time Surfaces (MCTS) which generalizes the commonly used time surface representation~\cite{Lag17} to an $n$-channel tensor of time surfaces for both polarities and different time window sizes. Since an optimal window size depends inversely on the motion magnitude, this approach further increases the model's robustness to both slow and fast motion. Additionally, we revisit network architecture choices of frame- and event-based vision to optimize our model's performance. Figure~\ref{pipeline} shows the complete data processing pipeline.\par
In absence of evaluation procedures to test event-based keypoint and descriptor stability, we develop a benchmark to assess the matched keypoint's quality for pose estimation, similar to benchmarks commonly used in frame-based vision~\cite{Sar20, Tys20, Sun21, Gle23}. On two datasets recorded with different sensors, settings, resolutions, and scenes, we outperform other data-driven and handcrafted approaches in terms of pose estimation by a large margin, demonstrating the robustness of our approach. We carry out ablation studies to confirm the effectiveness of our design choices.\par
Finally, we integrate our model into a state-of-the-art (SOTA) stereo visual inertial SLAM (VI-SLAM) framework~\cite{Leu22}, replacing the frame-based keypoint detector and descriptor extractor with SuperEvent. Our results surpass the current SOTA on event-based Visual Odometry and SLAM. This demonstrates the great potential of an event-sensor integration into existing, well-developed algorithms for frame-based vision.\par
We name our method \textit{SuperEvent} in honor of the great influence of the works SuperPoint~\cite{Det18} and SuperGlue~\cite{Sar20} on the research field of keypoint detection as well as on our model directly: ultimately, their frame-based detections and matches are used as pseudo-labels to teach keypoint detection and description to SuperEvent.\par
\textbf{Contributions}:
(1)~We propose a scalable method to generate training data for interest point detection and matching, leveraging robust existing frame-based keypoint detectors. The resulting data incorporates the real temporal dependence of event generation using real event camera recordings. This enables data-driven models to effectively learn robust keypoint detection and description.
(2)~We develop the model for event-based vision through key design choices: Our novel MCTS event representation increases the performance of data-driven models by reducing their dependence on motion speed of the scene. A transformer backbone shows to further improve the model's feature extraction. In our conducted experiments, our approach outperforms SOTA detection and matching by a large margin. We publicly release our code to support future event-based vision research. (3)~Showcasing the practicality of our method, we integrate the trained model into an existing stereo VI-SLAM approach developed for frame-based vision. The resulting event inertial SLAM system outperforms SOTA approaches on the commonly used TUM-VIE~\cite{Kle21} and VECtor~\cite{Gao22} datasets by a large margin. 
\section{Related Work}
\label{sec:related}

We begin with a short summary of keypoint detectors for frame-based vision since our work employs some of these findings for model design and training data generation. We then group the event-based works into handcrafted and data-driven keypoint detection, as well as tracking methods. Since our focus in this work lies on keypoint detection and matching based on event data exclusively, we do not consider approaches that rely on combination with other sensors, e.g., frame cameras. For an extensive overview of event-based vision, we refer the interested reader to~\cite{Gal20}.\par

\textbf{Frame-based keypoint detection:}
there exists a vast variety of approaches to detecting and describing keypoints in frames. Traditional, handcrafted methods~\cite{Har88, Shi94, Low04, Bay08, Cal10, Rub11, Leu11} are still widely used, since they are well studied and usually work robustly with little overhead. More recent, data-driven methods outperform the traditional keypoint detectors with impressive results. SuperPoint~\cite{Det18} is the first method trained with homographic adaptation, a self-supervised training technique of warping input images and their detected labels by randomized homographies. SuperPoint's architecture consists of a shared VGG~\cite{Sim14}-backbone with a detector and descriptor head, predicting descriptors in a discrete grid which is interpolated to full resolution. The authors later publish the extension SuperGlue~\cite{Sar20}, a graph neural network that improves the descriptor matching. Following works propose different architectures trained with homographic adaptation~\cite{Sun21, Tys20, Gle23}.\par

\textbf{Event-based keypoint detection: handcrafted approaches} can be broadly divided into filters and frame reconstruction. Filtering methods work directly on the event stream, attempting to find events originating from corners~\cite{Cla15, Alz18}. While their asynchronous nature offers little processing overhead, downstream tasks are required to process asynchronous data. Existing well-developed downstream solutions for frame-based vision therefore require major modifications.\par
Frame reconstruction methods collect events to construct a frame representation, such as binary event frames or time surfaces~\cite{Lag17}. Adopting methods from frame-based vision, such as the Harris corner detector~\cite{Har88} in~\cite{Vas16, Glo21} or the FAST detector~\cite{Ros06} in~\cite{Mue17}, these approaches detect corners in these event frames. However, they suffer from the inherent differences between reconstructed event frames and traditional frames, resulting in limited robustness of the algorithms developed for frame-based vision.\par
Additionally, these handcrafted methods usually employ simple nearest-neighbor matching in pixel space to track keypoints over time. This approach is only reliable for simple scenes since cluttered geometry, high variance in depth, and fast motions cause keypoints to become occluded, tightly clustered, or to jump. Also, the developed heuristics struggle with the ubiquitous and complex noise inherent to event cameras. Usually, they require manual parameter tuning for different camera models, datasets, or sometimes even scenes.\par

\textbf{Event-based keypoint detection: data-driven approaches} promise to cope better with the event data and its noise. A series of works~\cite{Man19, Chi21, Chi22} train data-driven corner detectors using labels by detecting keypoints on corresponding gray-scale frames that are synthetic or from an HVGA ATIS sensor. While these approaches learn data-driven detection, they only track keypoints by nearest-neighbor matching in pixel space, yielding the aforementioned drawback of limited robustness in complex scenes. Additionally, the employed corner detector for ground truth generation as well as the machine-learning models have limited performance due to their simplicity.\par
For these reasons, EventPoint~\cite{Hua23} and SD2Event~\cite{Gao24} employ more complex neural networks that additionally predict descriptors. While EventPoint uses the same architecture as SuperPoint~\cite{Det18} and finetunes its pre-trained weights, SD2Event employs agent-based attention to learn detection and description. Both approaches are trained with homographic adaptation and varying time window sizes. While this results in rotation-, distortion-, and scale-invariant models, the main challenge of keypoint description remains largely unsolved: Because most events result from scene or camera motion, corresponding keypoints vary in visual appearance. Since robust keypoint matching requires similar descriptor pairs, this motion dependence must be considered. However, homographic adaptation creates static frame pairs with little to no variance in scene motion within the samples. This contradicts the dynamic and motion-dependent nature of events, resulting in unstable keypoint matching between different timestamps.\par
Additionally, the authors of EventPoint propose the event representation Tencode, extending time surfaces with two additional channels to encode the events' polarities. However, to match SuperPoint's architecture with single-channel input, the Tencode tensor is then converted to one channel, disregarding the previously introduced separation of polarities.\par
Our method enforces robust descriptor matching by employing training data containing pseudo-labels for keypoints and descriptors under varying motion. We avoid synthetic event data since current simulators struggle to model realistic noise. Furthermore, our MCTS representation strictly separates the polarities and provides a range of time window sizes, providing additional information and omitting the time window parameter choice.

\textbf{Event-based keypoint tracking:} while early works fall back on batching events and processing events in a discretized way~\cite{Ni12, Kue16, Zhu17}, a promising idea is to exploit their spatial continuous visual flow~\cite{Alz18ace, Hu22, Dar23}. Among these approaches is HASTE~\cite{Alz20} which calculates the most likely motion hypothesis from a fixed set of translations and rotations for every incoming event. Since HASTE relies on external keypoint initialization, RATE~\cite{Iku24} employs a Shi-Tomasi~\cite{Shi94} corner detector on the binary event frame, initializing several HASTE instances to track multiple keypoints in parallel.\par
A major drawback of the described tracking approaches is their computational overhead when processing each event individually. Also, most methods only track one keypoint, resulting in an almost linear growth of computational cost with increasing number of tracker instances. E.g., RATE tracks only 20 features in real-time for an event stream of $180 \times 40$ pixels resolution -- which is hardly sufficient for downstream applications such as SLAM. Additionally, these handcrafted approaches struggle with environmental and hardware changes, as well as the event cameras' noise.\par
Our method SuperEvent works reliably in real-time with adjustable frequency and without re-training or excessive parameter tuning for different camera models. It's dense keypoint heatmap predictions barely result in overhead for more detections. In contrast to the tracking approaches, SuperEvent can be directly integrated into downstream algorithms relying on synchronous output and descriptors, e.g., for SLAM with place recognition, as we demonstrate.

\textbf{Event-based Odometry:} existing Event Odometry (EO) approaches are developed specifically for event processing. While some approaches combine events with frames~\cite{Vid18, Mah22, Che23, Gua23, Gua24_a}, event-only approaches can be classified as monocular EO~\cite{Reb16, Kle24}, monocular EO with IMU~\cite{Gua22, Gua24_b}, stereo EO~\cite{Zho21, Gho24}, and stereo EO with IMU~\cite{Liu23, Niu24_a, Niu24_b}. Due to the short history of event cameras, these systems require extensive research and development efforts to work reliably in practice.\par
In contrast, SuperEvent can be integrated into existing frame-based VO and SLAM systems. We thereby achieve event-based VI-SLAM leveraging highly accurate frame-based systems, emerging from years of research~\cite{Cam21, Leu22}.
\section{Method}
\label{sec:method}
In this section, we describe how we train SuperEvent, introduce the advantages and construction of the MCTS event representation, and detail the network design choices.

\subsection{Training Data Generation}~\label{sec:tdg}
To train SuperEvent, we exclusively employ real event camera recordings. Thus, our model learns to cope with the complex event stream properties, e.g., artifacts and noise, which are difficult to simulate realistically.\par
However, manually annotating keypoints in the asynchronous event data stream is very time-consuming and does not scale. Therefore, we propose an approach to generate ground truth data automatically. We take advantage of the pixel-synchronized events and frames of iniVation DAVIS\footnote{https://inivation.com/} event cameras. Employing a highly robust frame-based keypoint detector and matcher such as SuperPoint~\cite{Det18} + SuperGlue~\cite{Sar20} on the gray-scale frames, we can obtain temporally sparse pseudo-labels. To generate a diverse set of training data with reliable pseudo-labels, we process every frame $i$ of a sequence in the following way:
\begin{itemize}
    \item Firstly, we compute the distances in pixel-space of all key points to their respective matches in frame $i+1$. We only consider frames $i$ for training where the distances' median surpasses a threshold $c_\text{m}$. A small median distance indicates that the scene is nearly static, resulting in insufficient event data for a useful prediction.
    \item To expose the model to diverse camera pose changes and motion directions, we generate pseudo-labels by matching the descriptors with a subsequent frame $i+j$. We increase $j$ recursively, resulting in multiple pseudo-labels with increasing time interval between the frames:
    \begin{equation}
        j \leftarrow j + j_\text{s},\ j_\text{s} \sim \mathcal{U}(1, j_\text{max}),
    \end{equation}
    with the increment $j_\text{s}$ drawn from the uniform distribution $\mathcal{U}(1, j_\text{max})$. $j$ is initialized with 0 before its first update. The maximum step distance $j_\text{max}$ is manually adjusted depending on the properties and texture of the sequence, e.g., for sequences with fast motion, $j_\text{max}$ is decreased to yield sufficient matches.
    \item If the number of detected matches falls below the threshold $c_\text{m}$, these pseudo-labels are discarded, and the next frame $i+1$ becomes the new reference frame. This procedure ensures sufficient visual overlap and enough texture in the two frames.
\end{itemize}
Frames with under- or overexposure, or strong motion blur harm the prediction quality of the frame-based model. While these frames are usually filtered out since there are few matches (if $c_\text{d}$ is chosen adequately), we also manually remove scenes that mostly contain too dark, too bright, or blurred frames.

\subsection{Multi-channel Time Surface}

\begin{figure}
  \includegraphics[width=\columnwidth]{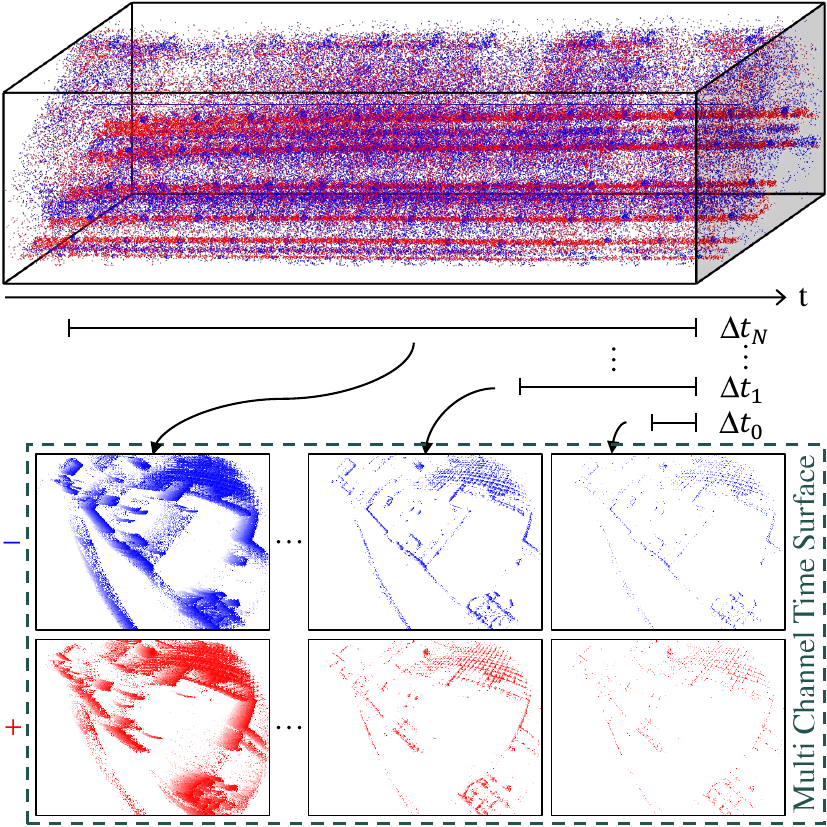}
  \caption{Schematic depiction of the MCTS generation process.}
  \label{mcts}
  \vspace{-10pt}
\end{figure}

To allow for efficient batch processing of event data, the asynchronous stream is converted to a tensor with fixed dimensions. In the literature, various variants have been proposed to encode the event stream. An ablation study in \cite{Mes23} compares some of the most commonly used event representations: binary event frames, time surfaces~\cite{Lag17}, and voxel grids~\cite{Zhu18_voxel}. For their data-driven feature tracking approach, time surfaces yield the best model performance. The time surface variant \textit{Tencode} with two more channels encoding the event polarity is proposed in \cite{Hua23}. It is used as model input after conversion to a single-channel frame.\par
Time surfaces and its variant Tencode have proven to be an efficient input representation for keypoint detection and tracking algorithms \cite{Mes23, Hua23, Gao24}, since they assign the highest values to the pixel locations of the most recent events that are usually caused by visual edges. This facilitates finding keypoints along these edges. However, these encodings have shortcomings:
\begin{itemize}
    \item Time surfaces are one-channel frames that ignore the event polarity and thereby discard potentially useful information. The Tencode representation is an attempt to overcome this issue. However, since it is converted back to a single channel tensor, there is still no clear separation between positive and negative events.
    \item Time surfaces and the Tencode representation depend on the time window length $\Delta t$. Increasing this parameter not only results in more events being considered but also decreases the contrast between the most recent events and past events. For high optical flow, this negatively impacts the detection of edges. Conversely, a smaller time window increases this contrast but includes fewer events, which can fail to sufficiently capture edges in areas with little optical flow.  Choosing a suitable $\Delta t$ is challenging, as it depends on the camera and scene motion which might be unknown \textit{a priori}. 
    The bottom part of Figure~\ref{mcts} shows the effect of varying $\Delta t$: Only the time surfaces in the middle clearly display the net on the top right.
\end{itemize}
Our proposed MCTS representation mitigates these problems: It combines time surfaces for each polarity and for various time windows of temporal size $\Delta t_n , \ n \in \{1,\ldots,N\}$. We distribute the time window sizes logarithmically to trade off data preprocessing and coverage of a wide variety of visual motion speeds. Events are encoded by their timestamp $t$, their pixel coordinates $x$ and $y$, and their polarity $p$. Mathematically, the MCTS tensor at time $\tau$ considering all $i$ events $\mathbf{e}_i = (t_i, x_i, y_i, p_i)$ between time $\tau-\Delta t_n$ and $\tau$ is defined as
\begin{multline}
    \text{\textbf{MCTS}} =
    (\text{\textbf{TS}}_{{-1}, \Delta t_1}, \ldots, \text{\textbf{TS}}_{{-1}, \Delta t_N,},\\ \text{\textbf{TS}}_{{+1}, \Delta t_1}, \ldots, \text{\textbf{TS}}_{{+1}, \Delta t_N }),
\end{multline}
with each time surface $\textbf{TS}$, initialized with zeros, defined by polarity $p \in \{-1, +1\}$ and $\Delta t_n$
\begin{equation}
   \text{\textbf{TS}}_{{p}, \Delta t_n}(x_i, y_i) = \max_{i, \ p_i=p}\left(1 - \frac{\tau - t_i}{\Delta t_n}\right).  
\end{equation}
The MCTS generation process is visualized in Figure~\ref{mcts}.

\subsection{Network Architecture \& Loss}
\label{architecture}

As commonly employed in frame-based keypoint detection models~\cite{Det18, Tys20, Gle23}, we combine a shared backbone with a detector and a descriptor head. Starting with the basic SuperPoint~\cite{Det18} architecture with adjusted loss functions for our temporal matching with pseudo-labels, we train various models with different backbones and hyperparameters and compare their performance. As an alternative approach, we replace the grid-based descriptor head with full resolution prediction and the respective loss functions as in~\cite{Tys20, Gle23}. Please refer to the full results of this ablation study in the ~\textbf{Supplementary Material}.\par
\begin{figure}[t]
  \includegraphics[width=\columnwidth]{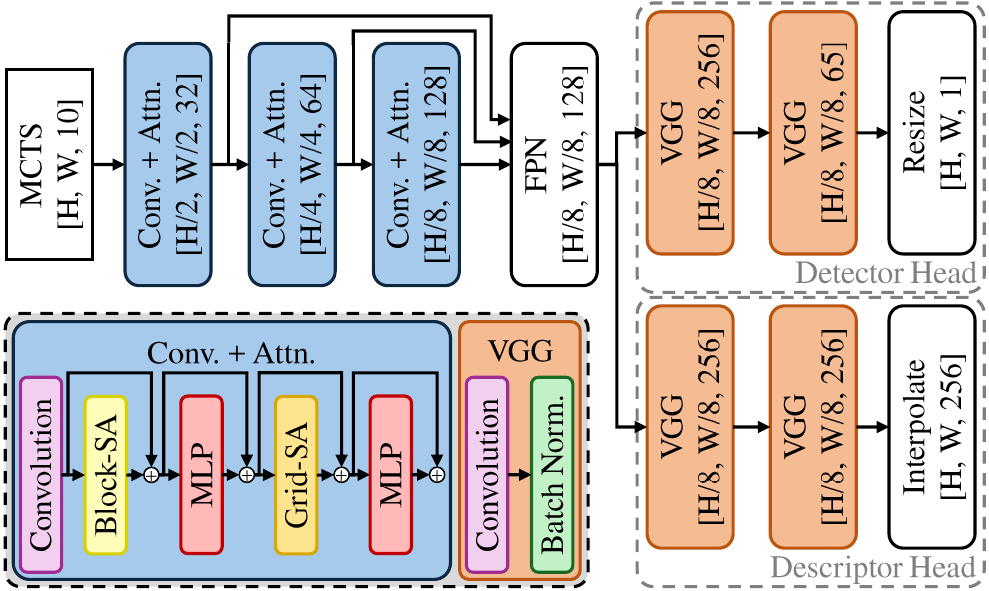}
  \caption{SuperEvent network architecture and tensor dimensions: a shared transformer backbone is combined with a detector and a descriptor head. The components of the Convolution-Attention blocks (\textit{Conv. + Attn.}) and the VGG blocks are displayed on the bottom left. Activations are omitted for simplicity.}
  \label{network}
\end{figure}

Our resulting architecture consists of a 3-layer MaxViT~\cite{Tu22} backbone with a Feature Pyramid Network (FPN)~\cite{Lin17} which is also employed in a highly efficient, event-based detector~\cite{Geh23}. Combined with grid-based VGG~\cite{Sim14} detector and descriptor heads, SuperEvents fully convolutional components can process various input resolutions without retraining. Figure~\ref{network} shows the final network architecture.\par
The loss function evaluates the network prediction for two corresponding training tensors $\{0, 1\}$. It combines the individual detector losses $L_{\text{s},0}$ and $L_{\text{s},1}$ with the joint descriptor loss $L_{\text{d}}$ weighted with the constant $c_\lambda$ 
\begin{equation}
    L = L_{\text{s},0} + L_{\text{s},1} + c_\lambda L_{\text{d}}.
\end{equation}

Following \cite{Det18}, the detection is implemented as a classification problem of the pixels in image patches, i.e., the pixel indices represent the ``classes''. The detector loss $L_{\text{CE}}$ is thus defined as cross-entropy between a predicted patch of keypoint scores $\mathbf{y}_{\text{s}_{h,w}}$ (dimension $8 \times 8$ plus no-keypoint-dustbin, thus 65) per grid cell (indices $h, w$) and the pseudo-labels $\hat{\mathbf{y}}_{\text{s}_{h,w}}$ in one-hot encoding
\begin{equation}
    L_{\text{s}} = \frac{1}{H_{\text{c}} W_{\text{c}}} \sum_{\substack{h=1\\w=1}}^{H_{\text{c}}, W_{\text{c}}} L_{\text{CE}}(\mathbf{y}_{\text{s}_{h,w}}, \hat{\mathbf{y}}_{\text{s}_{h,w}}),
\end{equation}
for $H_{\text{c}} \times W_{\text{c}}$ grid cells. In the case of multiple keypoint labels in a grid cell, one is randomly selected.\par

Unlike other models trained with homographic adaptation, our sparse pseudo-labels do not allow for dense descriptor training for every pixel or grid cell. As in \cite{Det18}, we only consider a single descriptor (256 channels) per grid cell, ${\mathbf{y}}_{\text{d}_{h_0,w_0}}$ and ${\mathbf{y}}_{\text{d}_{h_1,w_1}}$ (normalized to length one across the channels) with $(h_0, w_0)$ and $(h_1, w_1)$ denoting the grid indices of tensor 0 and 1. Successful keypoint matching requires similar descriptors for correspondences and distinct descriptors for unrelated keypoints. We measure descriptor similarity with the dot product $d^{h_0,w_0}_{h_1,w_1} ={\mathbf{y}}_{\text{d}_{h_0,w_0}} \cdot {\mathbf{y}}_{\text{d}_{h_1,w_1}}$. Our pseudo label $\hat{y}^{h_0,w_0}_{h_1,w_1} =1$ indicates that two descriptors at the respective grid cell locations correspond, otherwise $\hat{y}^{h_0,w_0}_{h_1,w_1} = 0$. Since the majority of the descriptors do not correspond, the partial loss of descriptor matches is weighted with the constant $c_{\text{d}}$. Mathematically, we formulate the descriptor loss as
\begin{equation}
    L_{\text{d}} = \sum_{\substack{h_0=1\\w_0=1}}^{H_{\text{c}}, W_{\text{c}}} \sum_{\substack{h_1=1\\w_1=1}}^{H_{\text{c}}, W_{\text{c}}}
    \begin{cases}
        c_\text{d} \text{max}( 0, c_\text{p} - d^{h_0,w_0}_{h_1,w_1}), &\hat{y}^{h_0,w_0}_{h_1,w_1}=1, \\
        \hfill
        \text{max}( 0, d^{h_0,w_0}_{h_1,w_1} - c_\text{n} ), &\hat{y}^{h_0,w_0}_{h_1,w_1}=0.
    \end{cases}
\end{equation}
The constants $c_{\text{p}}$ and $c_{\text{n}}$ lead to saturation for maximal desired descriptor (dis-)similarity. All descriptors of grid cells without a keypoint label index (${\mathbf{y}}_{\text{d}_{h_0,w_0}} = 0$ or ${\mathbf{y}}_{\text{d}_{h_1,w_1}} = 0$) do not contribute to the loss.\par

During inference, we employ a simplified, yet more effective Non-Minimum-Suppression (NMS)~\cite{Neu06} approach in the context of keypoint detection. Instead of the common approach to model keypoints as fixed-size boxes and then apply classical NMS as in~\cite{Det18, Hua23}, we find local maxima in the predicted keypoint score heatmap. In contrast to classical NMS, our method only considers local maxima and not all keypoint candidates sufficiently far from other points with higher scores. Additionally, it employs parallelizable operations instead of consecutive updates of the keypoints sorted by score. Every predicted score $s$ at the pixel coordinates $(u,v)$ is considered for a potential keypoint if
\begin{multline}
    s[u,v] > s[u+\Delta u, v+\Delta v]\\
    \forall \Delta u,\Delta v \in [-c_{\text{b}}, \ldots, -1, 1, \ldots, c_{\text{b}}].
\end{multline}
with $c_{\text{b}}$ determining the size of the local neighborhood. The keypoint candidates are further filtered by a minimal required score $c_\text{s}$, or the $n_\text{s}$ highest scores are selected.

\section{Experiments}
\label{sec:experiments}

\begin{table}[t]
\begin{center}
\caption{Inference times of our compiled PyTorch model on GPU for common event-stream resolutions (cropped to compatible shape) using float16 precision (same accuracy as float32 in our experiments). NMS requires an additional \SI{0.4}{ms} on average.\label{runtime}}
\small
\begin{tabularx}{\linewidth}{
>{\centering}X
>{\centering}X
>{\centering}X
>{\centering\arraybackslash}X
}
\toprule
$160 \times 240$ & $240 \times 320$ & $480 \times 640$ & $720 \times 1280$ \\
\midrule
\SI{1.3}{ms} & \SI{2.6}{ms} & \SI{7.8}{ms} & \SI{23.2}{ms} \\
\bottomrule
\end{tabularx}
\end{center}
\end{table}

To validate the effectiveness of the presented approach, we conduct a series of experiments and ablation studies. Firstly, we evaluate the detection and description of keypoints by estimating the camera pose changes based on the matched points. We outperform other keypoint detectors and trackers~\cite{Chi22, Hua23, Iku24} by a large margin.\par
In our second experiment, we integrate SuperEvent into the SLAM framework OKVIS2~\cite{Leu22} originally developed for frame-based cameras. We achieve convincing results on the widely used event-SLAM benchmarks~\cite{Kle21, Gao22} without extensive tuning, showcasing SuperEvent's versatility.\par
For ablation studies, more qualitative examples, a detailed description of the pose estimation benchmark, and a comparison to frame-based matching for high-speed/ HDR scenarios, please refer to our \textbf{Supplementary Material}.

\subsection{Implementation Details}
We generated training data as introduced in Section~\ref{sec:tdg}. To expose the model to various scenes and camera intrinsics, we combine multiple datasets~\cite{Zhu18, Del19, Hu20, Rod21, Lee22}, obtaining more than 100k training sample pairs with pseudo-labels. We implement SuperEvent in PyTorch~\cite{Ans24} and train it with the Adam~\cite{Kin14} optimizer (learning rate $1e^{-4}$, betas $0.9, 0.999$) for 10 epochs.\par
We employ the network architecture as in Section~\ref{architecture} with the loss function hyperparameters $c_\lambda = 10$, $c_{\text{d}} = 0.5$, $c_{\text{p}} = 1.0$, and $c_{\text{n}} = 0.2$. We encode events as a 10-channel MCTS with 5 time window sizes $\Delta t$ logarithmically distributed from \SI{0.001}{s} to \SI{0.1}{s}. The detection threshold is set to $c_{\text{s}} = 0.01$ and the NMS neighborhood $c_{\text{b}} = 2$ pixels.\par
We evaluate on a desktop PC with an Intel Core i5-13600 processor, an NVIDIA GeForce GTX 4070 GPU, and \SI{32}{GB} RAM. Inference times are shown in Table~\ref{runtime}.

\subsection{Pose Estimation}

\begin{table}[t]
\begin{center}
\caption{Pose estimation on the Event Camera dataset.\label{ecd_pose_est}}
\small
\begin{tabularx}{\linewidth}{
l
>{\centering}X
>{\centering}X
>{\centering\arraybackslash}X
}
\toprule
\multirow{2}{*}{Method} & \multicolumn{3}{c}{ECD: Pose Estimation AUC in \%} \\
\cmidrule(lr){2-4}
 & @5° & @10° & @20° \\
\midrule
LLAK~\cite{Chi22} & 0.7 & 1.4 & 2.1 \\

RATE~\cite{Iku24} & \underline{3.3} & \underline{8.4} & \underline{18.0} \\

EventPoint~\cite{Hua23} & 1.6 & 3.0 & 5.4 \\

\textbf{SuperEvent (ours)} & \textbf{22.7} & \textbf{35.8} & \textbf{46.7} \\
\bottomrule
\vspace{-28pt}
\end{tabularx}
\end{center}
\end{table}

A common benchmark for evaluating frame-based keypoint detectors is pose estimation~\cite{Sar20, Tys20, Sun21, Gle23}. These works use datasets such as ScanNet~\cite{Dai17} and MegaDepth~\cite{Li18} containing various images of the same scene with the associated ground truth camera poses. To create a similar event-based benchmark, we employ sequences with ground truth poses from real-world dynamic datasets. We search for uniformly distributed pose changes between 1° and 45°. The upper bound is chosen to increase the probability of visual overlap. However, since we do not know the ground truth geometry of the scene, it cannot be guaranteed that every sample contains sufficient information to recover the ground truth pose. Next, we process the event stream with various approaches to detect and match keypoints. After outlier removal with Random Sample Consensus (RANSAC)~\cite{Fis81}, we estimate the camera pose and calculate the error with respect to the ground truth. Finally, we report the area-under-curve (AUC) metric for different thresholds.\par
We compare the pose estimation capability of SuperEvent to the 3 approaches LLAK~\cite{Chi22}, RATE~\cite{Iku24}, and EventPoint~\cite{Hua23}. While we use the open-source implementations of LLAK and RATE, we re-implemented and trained EventPoint by carefully following the steps outlined in the paper. We were unable to evaluate the approach SD2Event~\cite{Gao24} due to unpublished code and missing implementation details in the respective work.\par
The experiment is conducted on two diverse datasets: Firstly, the Event Camera Dataset (ECD)~\cite{Mue17_ecd} is widely used to evaluate event-based approaches, as it was one of the first publicly available datasets. It is challenging due to the low resolution ($180 \times 240$), the noisy output of the iniVation DAVIS240C event camera, and fast motion changes. Secondly, we use the more recent dataset Event-aided Direct Sparse Odometry (EDS)~\cite{Hid22}. In contrast to our training data, this dataset is recorded with a Prophesee Gen 3.1\footnote{https://www.prophesee.ai/} event camera with a higher resolution of $480 \times 640$ pixels and an over-proportionally increased event rate. Fast, inconsistent camera motions and a high dynamic range make pose estimation challenging.\par
Table~\ref{ecd_pose_est} shows that SuperEvent outperforms all baselines by a large margin. On the ECD dataset, its advantage is the greatest for the high-precision estimations below 5° error, confirming the effectiveness of our method. In contrast, we found that LLAK detects many keypoints but cannot track the vast majority of them. EventPoint's matching capability is severely limited, possibly due to their static assumptions during training inherently contradicting the dynamic nature of event cameras. Finally, RATE achieves decent results for estimations with up to 20° error, but since it cannot recover from tracking loss, it depends on consistent and precise tracking of most keypoints. As seen in the results, this requirement cannot be stably fulfilled.\par

\begin{table}[t]
\begin{center}
\caption{Pose estimation on Event-aided Direct Sparse Odometry.\label{eds_pose_est}}
\small
\begin{tabularx}{\linewidth}{
l
>{\centering}X
>{\centering}X
>{\centering\arraybackslash}X
}
\toprule
\multirow{2}{*}{Method} & \multicolumn{3}{c}{EDS: Pose Estimation AUC in \%} \\
\cmidrule(lr){2-4}
 & @5° & @10° & @20° \\
\midrule
LLAK~\cite{Chi22} & 0.5 & 0.7 & 1.0 \\

RATE~\cite{Iku24} & \underline{2.1} & \underline{5.1} & \underline{10.3} \\

EventPoint~\cite{Hua23} & 1.6 & 2.8 & 5.2 \\

\textbf{SuperEvent (ours)} & \textbf{15.2} & \textbf{26.4} & \textbf{40.1} \\
\bottomrule
\vspace{-28pt}
\end{tabularx}
\end{center}
\end{table}

Table~\ref{eds_pose_est} indicates a slight decrease in SuperEvent's pose estimation performance on the EDS dataset compared to ECD. Since all approaches perform worse, we attribute this effect to the dataset being more challenging. Interestingly, the performance of SuperEvent for 20° precision is still in the same region as for the ECD dataset. Considering the different properties of the Prophesee camera compared to the DAVIS camera employed in the training sequences, these results confirm the generalization capability of SuperEvent.\par

\subsection{Stereo Event-Visual Inertial SLAM}

\begin{table*}[t]
\begin{center}
\caption{Results on the small scale \textit{mocap} sequences of TUM-VIE dataset, ATE in cm. The first three gray results require scale alignment with the ground truth; all other approaches estimate the absolute scale. Baseline numbers taken from~\cite{Niu24_b, Gua24_b}.\label{tum_eo}}
\small
\begin{tabular*}{\linewidth}{@{\extracolsep{\fill}} =l +l +c +c +c +c +c +c}
\toprule
Method & Modality & 1d-trans & 3d-trans & 6dof & desk & desk2 & Average \\
\midrule
\rowstyle{\color{gray}}
EVO~\cite{Reb16} w/ scale alignment & Mono E & 7.50 & 12.50 & 85.50 & 54.10 & 75.20 & 46.96 \\
\rowstyle{\color{gray}}
DEVO~\cite{Kle24} w/ scale alignment & Mono E & 0.50 & \textbf{1.10} & 1.60 & 1.70 & 1.00 & 1.18 \\
\rowstyle{\color{gray}}
DEIO~\cite{Gua24_b} w/ scale alignment & Mono E + IMU & \textbf{0.42} & 1.11 & \textbf{1.37} & \textbf{1.36} & \textbf{0.73} & \textbf{1.00} \\
\midrule
DEIO~\cite{Gua24_b} & Mono E + IMU & 1.08 & \underline{1.12} & \underline{1.39} & \underline{1.41} & \underline{1.19} & \underline{1.24} \\

ESVO~\cite{Zho21} & Stereo E & 12.54 & 17.19 & 13.46 & 12.92 & 4.42 & 12.11 \\

ES-PTAM~\cite{Gho24} & Stereo E & \underline{1.05} & 8.53 & 10.25 & 2.50 & 7.20 & 5.91 \\

ICRA'24~\cite{Niu24_a} & Stereo E + IMU & 3.85 & 18.90 & failed & 8.99 & 9.47 & \text{--} \\

ESVO2~\cite{Niu24_b} & Stereo E + IMU & 3.33 & 7.26 & 3.21 & 6.16 & 4.02 & 4.78 \\

\textbf{OKVIS2~\cite{Leu22} + SuperEvent (ours)} & Stereo E + IMU & \textbf{0.44} & \textbf{0.89} & \textbf{0.43} & \textbf{0.58} & \textbf{0.41} & \textbf{0.55} \\

\textbf{(ours without loop closure)} & & (0.43) & (0.89) & (0.43) & (0.70) & (0.40) & (0.57) \\

\bottomrule
\vspace{-22pt}
\end{tabular*}
\end{center}
\end{table*}


\begin{table}[t]
\begin{center}
\caption{Results on VECtor~\cite{Gao22} large scale sequences. Our results are obtained with online calibration. Baseline numbers are taken from \cite{Gua24_b} and \cite{Kle24}. DEVO~\cite{Kle24} cannot recover the absolute scale. Sequences marked '--' are not evaluated in~\cite{Gua24_b}. ATE in cm.\label{vector}}
\small
\tabcolsep=0.0pt
\begin{tabularx}{\linewidth}{
l
>{\centering}X
>{\centering}X
>{\centering}X
>{\centering}X
>{\centering}X
>{\centering\arraybackslash}X
}
\toprule
 & corr.- & corr.- & units- & units- & school- & school- \\
Method & dolly & walk & dolly & scooter & dolly & scooter \\
\midrule

\textbf{Ours} & \textbf{33.13} & \textbf{133.16} & \textbf{122.61} & \textbf{59.05} & \textbf{69.96} & \textbf{39.07} \\

DEIO~\cite{Gua24_b} & \underline{492.65} & \underline{325.00} & \underline{826.38} & \underline{304.14} & \text{--} & \text{--} \\
ESVO~\cite{Zho21} & failed & failed & failed & failed & \underline{1371.0} & \underline{983.0} \\
\bottomrule
\end{tabularx}
\vspace{-18pt}
\end{center}
\end{table}

We demonstrate the usefulness of SuperEvent with a downstream experiment. To date, frame-based visual odometry and SLAM systems largely outperform their event-based counterparts, attributed to their decades-long lead in development. By integrating SuperEvent into the frame-based VI-SLAM system OKVIS2~\cite{Leu22}, we enable this framework to process the keypoints detected in the event stream. We compare the performance of the resulting event-only system to other recent event-based Odometry and SLAM approaches on the TUM-VIE dataset~\cite{Kle21} in Table~\ref{tum_eo}. Moreover, note that the next best results of the data-driven Mono E(I)O systems DEVO and DEIO are achieved only after aligning the scale of the estimated trajectories with ground truth. However, real-world applications, e.g., VR and robotics, typically require accurate metric scale. Results on the VECtor dataset~\cite{Gao22} are listed in Table~\ref{vector}.\par

OKVIS2's loop closure capability relies on long-term descriptor matching. While its effect on the small-scale sequences is limited, we evaluate OKVIS2 + SuperEvent on the \textit{loop-floor} sequences of the TUM-VIE dataset, since these are explicitly designed to test a system's loop-closure capabilities. Thanks to SuperEvent, the OKVIS2 loop-closure feature works reliably with events, as shown in Table~\ref{loop_floor}. Figure~\ref{loop_floor0} qualitatively compares the estimated trajectories of the sequence \textit{loop-floor0}. Without loop closure, there is significant drift -- albeit not out of the ordinary compared to frame-based VI-SLAM. However, with loop closure enabled, the starting room is recognized and the drift is corrected. This demonstrates the robustness of SuperEvent, achieving successful descriptor matching even with minutes of time between observations.\par

\afterpage{
\begin{table}[t]
\begin{center}
\caption{Effect of loop closure of OKVIS2~\cite{Leu22} + SuperEvent on TUM-VIE \textit{loop-floor} sequences, ATE in cm.\label{loop_floor}}
\small
\begin{tabularx}{\linewidth}{lcccc}
\toprule
loop-floor & 0 & 1 & 2 & 3 \\

Estimated length & \SI{349}{m} & \SI{316}{m} & \SI{279}{m} & \SI{303}{m} \\
\midrule
Loop closure & \textbf{4.96} & \textbf{4.64} & \textbf{8.92} & \textbf{4.74} \\

W/o loop closure & 132.11 & 161.92 & 116.00 & 129.17 \\
\bottomrule
\vspace{-24pt}
\end{tabularx}
\end{center}
\end{table}

\begin{figure}[t]
\includegraphics[width=\columnwidth]{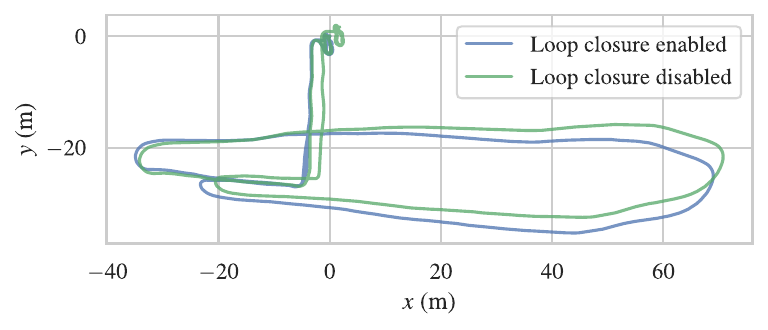}
\vspace{-20pt}
\caption{OKVIS2 + SuperEvent's estimated trajectories of TUM-VIE sequence \textit{loop-floor0} with and without loop closure.\label{loop_floor0}}
\vspace{-10pt}
\end{figure}
}

\section{Conclusion}
We presented SuperEvent, a novel keypoint detector and descriptor for event streams. During its training, we consider the time-dependent keypoint appearance and employ pseudo-labels detected on frames. Combined with the flexible MCTS representation and a transformer backbone, it achieves SOTA performance on keypoint-based pose estimation. SuperEvent's integration into frame-based downstream applications enables event vision in existing keypoint-based systems.\par
As future work, SuperEvent can be combined with a frame-based detector to realize systems that simultaneously process frames and events, exploiting their complementary advantages. Also, with a similar approach to SuperEvent, an event-based line detector could improve the performance of downstream applications in man-made environments.\par
Finally, our training data is generated to enable consistent feature description independent of their motion-dependent appearance. However, since we employ real sequences, consecutive samples are biased to having similar motion directions. This is reflected in the performance of SuperEvent, whose descriptor matchability suffers under strong motion changes. While this is an inherent property of event cameras, a different approach to generating training data could mitigate it.
{
    \small
    \bibliographystyle{ieeenat_fullname}
    \bibliography{main}
}

\newpage
\appendix
\twocolumn[
\begin{@twocolumnfalse}
    {\centering\section*{\Large Supplementary Material~\label{supp}}}
    \vspace{20pt}
\end{@twocolumnfalse}
]
\section{Ablation Studies}
To confirm the effectiveness of our approach, we conduct ablation studies regarding the training data generation process \textit{temporal matching}, and our proposed event representation MCTS. Additionally, we report the performance of our investigated network architectures.

\subsection{Training Data}

\begin{table*}[bp]
\vspace{24pt}
\begin{center}
\caption{Pose estimation after training SuperEvent with temporal matching data, homographic adaptation data, and samples from both methods.\label{train_abl}}
\begin{tabularx}{\linewidth}{
l
>{\centering}X
>{\centering}X
>{\centering}X
>{\centering}X
>{\centering}X
>{\centering\arraybackslash}X
}
\toprule
& \multicolumn{6}{c}{Pose Estimation AUC in \%} \\
\cmidrule(lr){2-7}
Training Data Generation Method & \multicolumn{3}{c}{Event Camera Dataset~\cite{Mue17_ecd}} & \multicolumn{3}{c}{Event-aided Direct Sparse Odm.\cite{Hid22}} \\
\cmidrule(lr){2-4}
\cmidrule(lr){5-7}
& @5° & @10° & @20° & @5° & @10° & @20° \\
\midrule
Homographic adaptation & 17.2 & 24.3 & 31.0 & 12.3 & 21.0 & 31.5 \\

\textbf{Temporal matching (ours)} & \textbf{22.7} & \textbf{35.8} & \textbf{46.7} & \textbf{15.2} & \textbf{26.4} & \textbf{40.1} \\

Homographic adaptation + temp. matching & \underline{18.5} & \underline{28.1} & \underline{37.1} & \underline{13.1} & \underline{22.0} & \underline{33.0} \\
\bottomrule
\end{tabularx}
\end{center}
\vspace{12pt}
\end{table*}

\begin{table*}[bp]
\begin{center}
\caption{Pose estimation with different time surface variants as input representations. The index number of the Multi Channel Time Surfaces (MCTS) indicates the number of time window sizes $\Delta t$.\label{input_abl}}
\begin{tabularx}{\linewidth}{
l
>{\centering}X
>{\centering}X
>{\centering}X
>{\centering}X
>{\centering}X
>{\centering}X
>{\centering\arraybackslash}X
}
\toprule
\multicolumn{2}{c}{\multirow{2}{*}{Input}} & \multicolumn{6}{c}{Pose Estimation AUC in \%} \\
\cmidrule(lr){3-8}
& & \multicolumn{3}{c}{Event Camera Dataset~\cite{Mue17_ecd}} & \multicolumn{3}{c}{Event-aided Direct Sparse Odometry~\cite{Hid22}} \\
\cmidrule(lr){1-2}
\cmidrule(lr){3-5}
\cmidrule(lr){6-8}
Representation & Channels & @5° & @10° & @20° & @5° & @10° & @20° \\
\midrule
Time Surface~\cite{Lag17} & 1 & 13.8 & 21.4 & 29.3 & 13.0 & 22.5 & 34.1 \\

Tencode (gray)~\cite{Hua23} & 1 & 19.5 & 28.7 & 36.9 & 13.9 & 23.7 & 36.3 \\

$\text{MCTS}_1$ (ours) & 2 & \underline{20.3} & \underline{30.0} & \underline{38.7} & \underline{14.1} & \underline{24.4} & \underline{37.0} \\

\textbf{$\text{MCTS}_5$ (ours)} & 10 & \textbf{22.7} & \textbf{35.8} & \textbf{46.7} & \textbf{15.2} & \textbf{26.4} & \textbf{40.1} \\
\bottomrule
\end{tabularx}
\end{center}
\end{table*}

We compare our proposed temporal matching approach with the state-of-the-art method homographic adaptation introduced for frame-based keypoint detection and description~\cite{Det18} in Table~\ref{train_abl}. SuperEvent trained with temporal matching data exclusively outperforms models trained with homographic adaptation data. Also, combining both approaches results does not improve model performance.\par
Temporal matching employs real data exclusively without distortions in the event representation due to augmentations. We suspect that this advantage improves the model's event data comprehension.

\subsection{Input Representation}

Next, we compare our MCTS representation to time surfaces~\cite{Lag17} and its variant Tencode~\cite{Hua23} in Table~\ref{input_abl}. As shown in~\cite{Hua23}, with $\Delta t = \SI{0.01}{s}$, the Tencode model achieves superior performance to the one using time surfaces. However, since Tencode is also used as a single channel tensor, an MCTS with a single time window size (but two channels) strictly separates polarities, thereby providing the model with additional information improving the performance further. Finally, more time windows increase the model's robustness to fast or slow scene motions. Therefore, the 10-channel $\text{MCTS}_5$ with $\Delta t_{\{1, \ldots, N\} } = \{\SI{0.001}{s}, \SI{0.003}{s}, \SI{0.01}{s}, \SI{0.03}{s}, \SI{0.1}{s}\}$ enables the model to outperform all other variants.

\subsection{Network Architecture}

\begin{table*}[t]
\begin{center}
\caption{
\centering
Network architecture ablation study on pose estimation on the Event Camera dataset~\cite{Mue17_ecd}. Every backbone layer reduces the spatial dimensions by half (except for $^3$).\\ $^1$Architecture similar to SuperPoint~\cite{Det18}\\
$^2$Architecture similar to DISK~\cite{Tys20}\\
$^3$Architecture similar to SiLK~\cite{Gle23} (no spatial reduction in backbone)\\
$^4$SuperEvent\\
$^5$Backbone similar to~\cite{Geh23}\\
$^6$Other investigated architectures
\label{net_arch}}
\begin{tabularx}{\linewidth}{
ll
>{\centering}X
>{\centering}X
>{\centering}X
>{\centering}X
>{\centering}X
>{\centering}X
>{\centering}X
>{\centering\arraybackslash}X
}
\toprule
& \multicolumn{3}{c}{Backbone} & Descriptor & \multicolumn{2}{c}{Loss} & \multicolumn{3}{c}{Pose Estimation AUC in \%} \\
\cmidrule(lr){2-4}
\cmidrule(lr){6-7}
\cmidrule(lr){8-10}
& Blocks & Layers & Channels & Resolution & Detector & Descriptor & @5° & @10° & @20° \\
\midrule
$^1$ & VGG & 3 & 32, 64, 128 & 8x8 grid & Cross-Entropy & dot product & 20.2 & 31.7 & 42.2 \\

$^2$ & VGG & 3 & 32, 64, 128 & pixelwise & Focal loss & Cycle-Consistency & 20.8 & 31.0 & 40.7 \\

$^3$ & VGG & 3 & 32, 64, 128 & pixelwise & Focal loss & Cycle-Consistency & 18.5 & 25.5 & 31.3 \\

$^4$ & MaxVit & 3 & 32, 64, 128 & 8x8 grid & Cross-Entropy & dot product & \textbf{22.7} & \textbf{35.8} & \textbf{46.7} \\

$^5$ & MaxVit & 4 & 32, 64, 128, 256 & 8x8 grid & Cross-Entropy & dot product & \underline{22.4} & \underline{33.9} & \underline{43.8} \\

$^6$ & MaxVit & 3 & 32, 64, 128 & pixelwise & Focal loss & Cycle-Consistency & 20.5 & 29.7 & 38.0 \\

$^6$ & MaxVit & 3 & 64, 128, 256 & pixelwise & Focal loss & Cycle-Consistency & 21.0 & 30.3 & 38.6 \\

$^6$ & MaxVit & 4 & 32, 64, 128, 256 & pixelwise & Focal loss & Cycle-Consistency & 20.3 & 30.5 & 40.3 \\

$^6$ & MaxVit & 4 & 64, 128, 256, 512 & pixelwise & Focal loss & Cycle-Consistency & 20.2 & 29.4 & 38.2 \\

$^6$ & MaxVit & 5 & 32, 64, 128, 256, 512 & pixelwise & Focal loss & Cycle-Consistency & 15.2 & 23.0 & 31.3 \\

$^6$ & MaxVit & 5 & 64, 128, 256, 512, 1024 & pixelwise & Focal loss & Cycle-Consistency & 17.5 & 26.4 & 35.6 \\
\bottomrule
\end{tabularx}
\end{center}
\end{table*}

We compare various combinations of network architectures from the literature~\cite{Det18, Gle23, Geh23}. We investigate two backbone configurations, namely VGG~\cite{Sim14} and MaxVit~\cite{Tu22}, and their hyperparameters \textit{number of layers in backbone} and \textit{output channels per layer in backbone}. Additionally, we investigated if a descriptor prediction on pixel-level as in~\cite{Tys20, Gle23} performs better than the 8-grid interpolation from~\cite{Det18}. For the pixelwise descriptor approach, we employ Focal loss~\cite{Lin17_b} to train the detector head and the Cycle-Consistency loss~\cite{Tys20, Gle23} for the descriptor head.
\newpage
\section{Examples of Network Predictions and Pseudo-labels}

\begin{figure*}[!ht]
    \centering
    \begin{subfigure}{0.49\textwidth}
        \centering
        \includegraphics[width=0.975\textwidth]{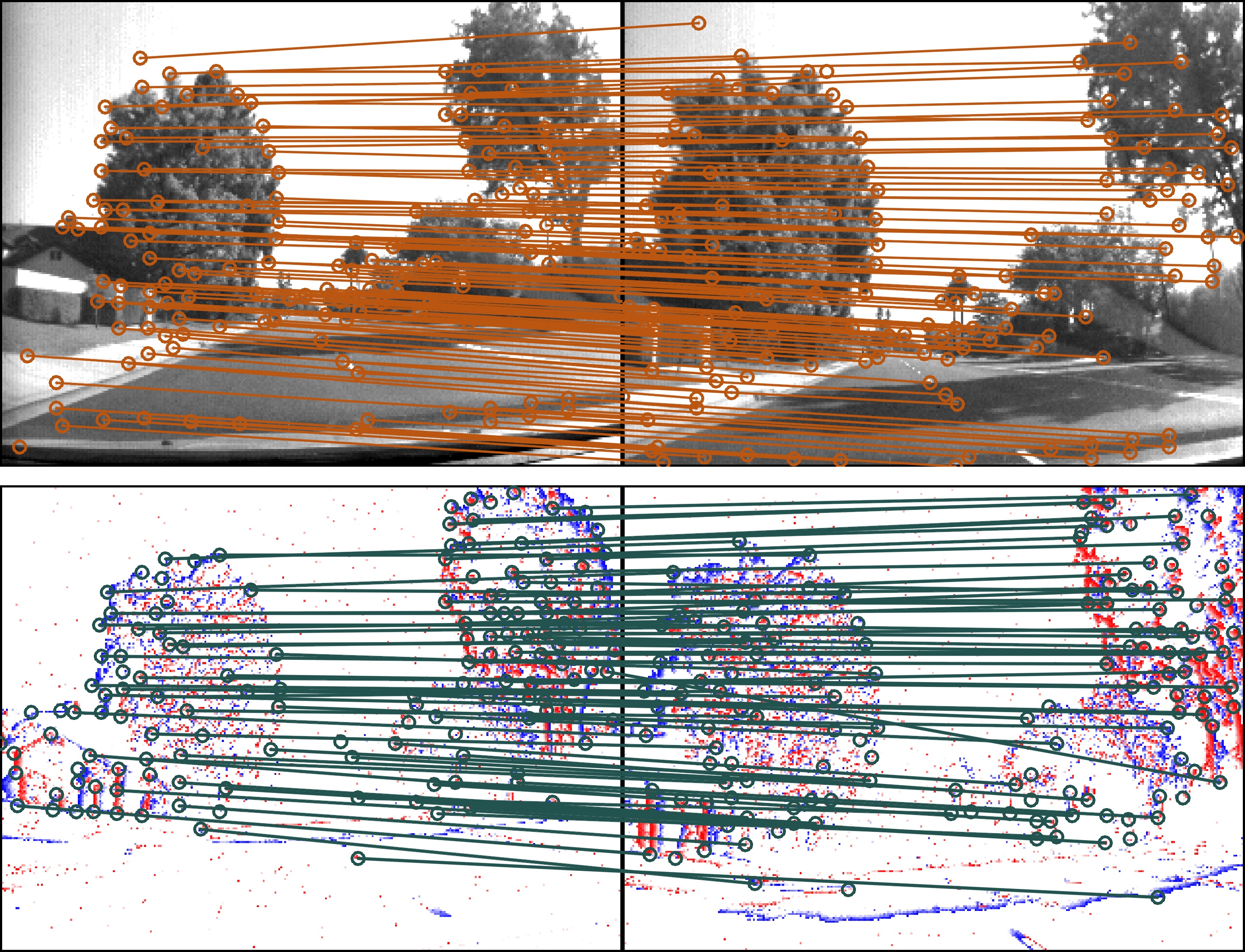}
        \caption{DAVIS Driving Dataset 2020~\cite{Hu20}: rec1501953155}
    \end{subfigure}
    \begin{subfigure}{0.49\textwidth}
        \centering
        \includegraphics[width=0.975\textwidth]{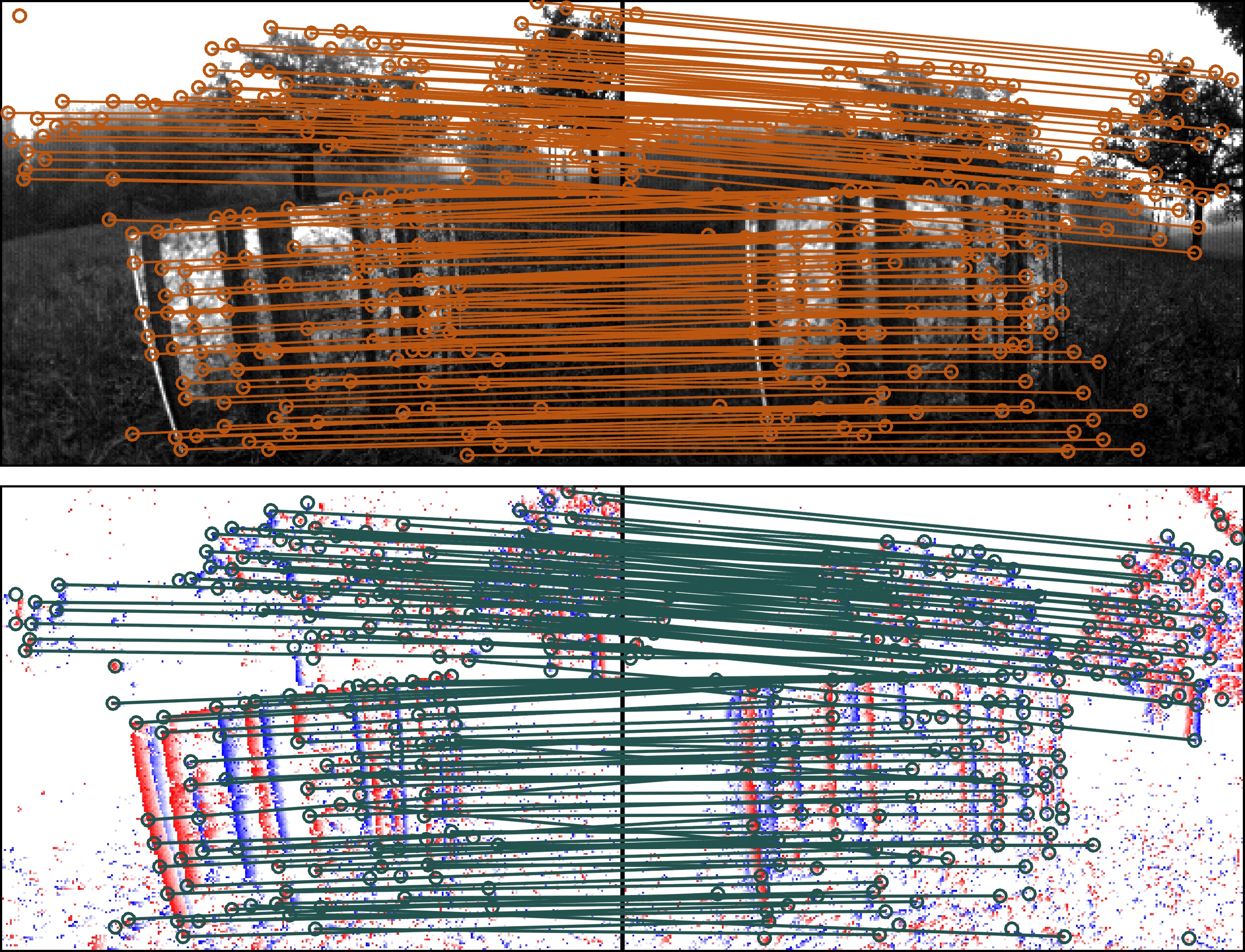}
        \caption{UZH-FPV Drone Racing Dataset~\cite{Del19}: Outdoor Forward Facing 2}
    \end{subfigure}\vspace{5mm}
    \begin{subfigure}{0.49\textwidth}
        \centering
        \includegraphics[width=0.975\textwidth]{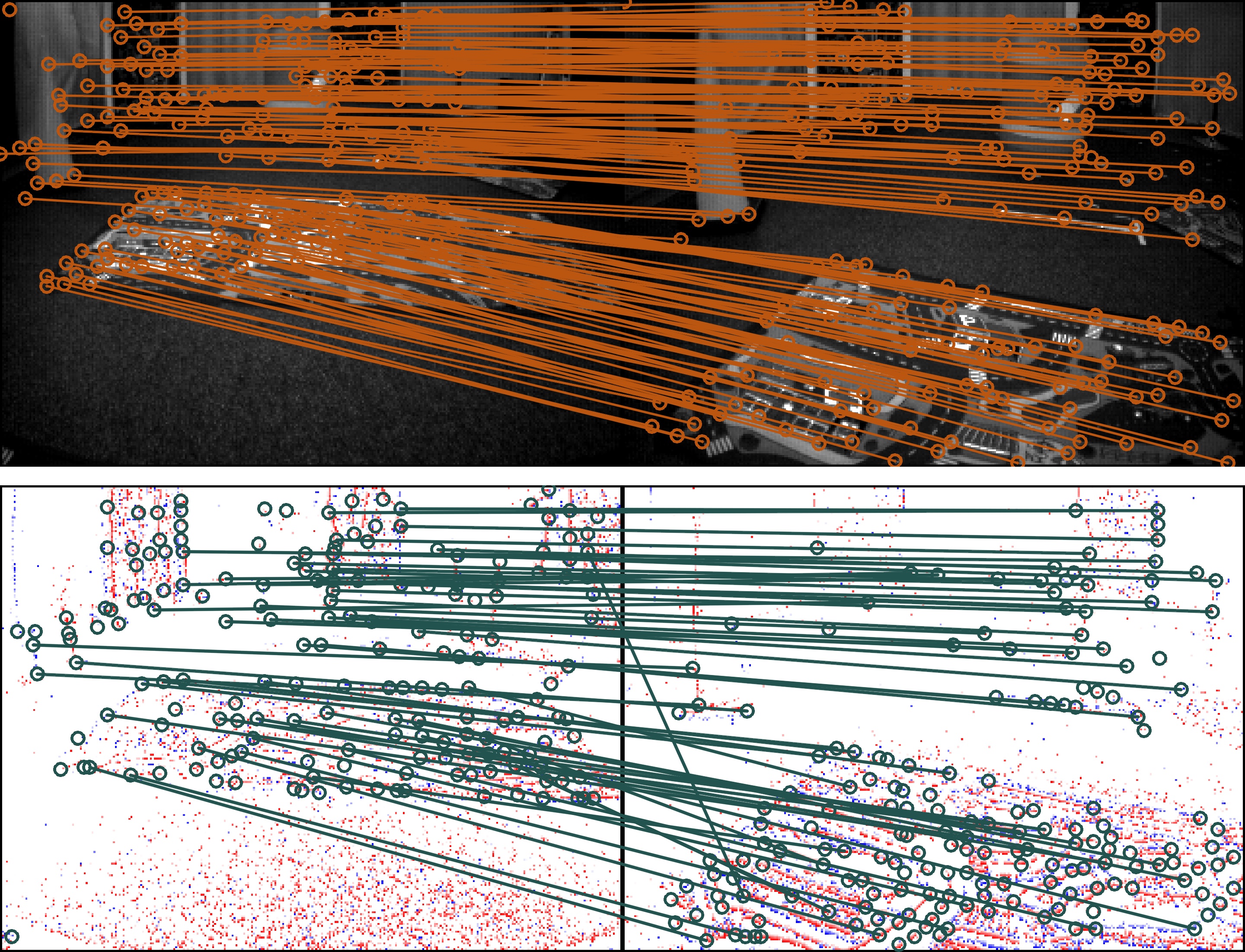}
        \caption{Multi Vehicle Stereo Event Camera Dataset~\cite{Zhu18}: Indoor Flying 2}
    \end{subfigure}
    \begin{subfigure}{0.49\textwidth}
        \centering
        \includegraphics[width=0.975\textwidth]{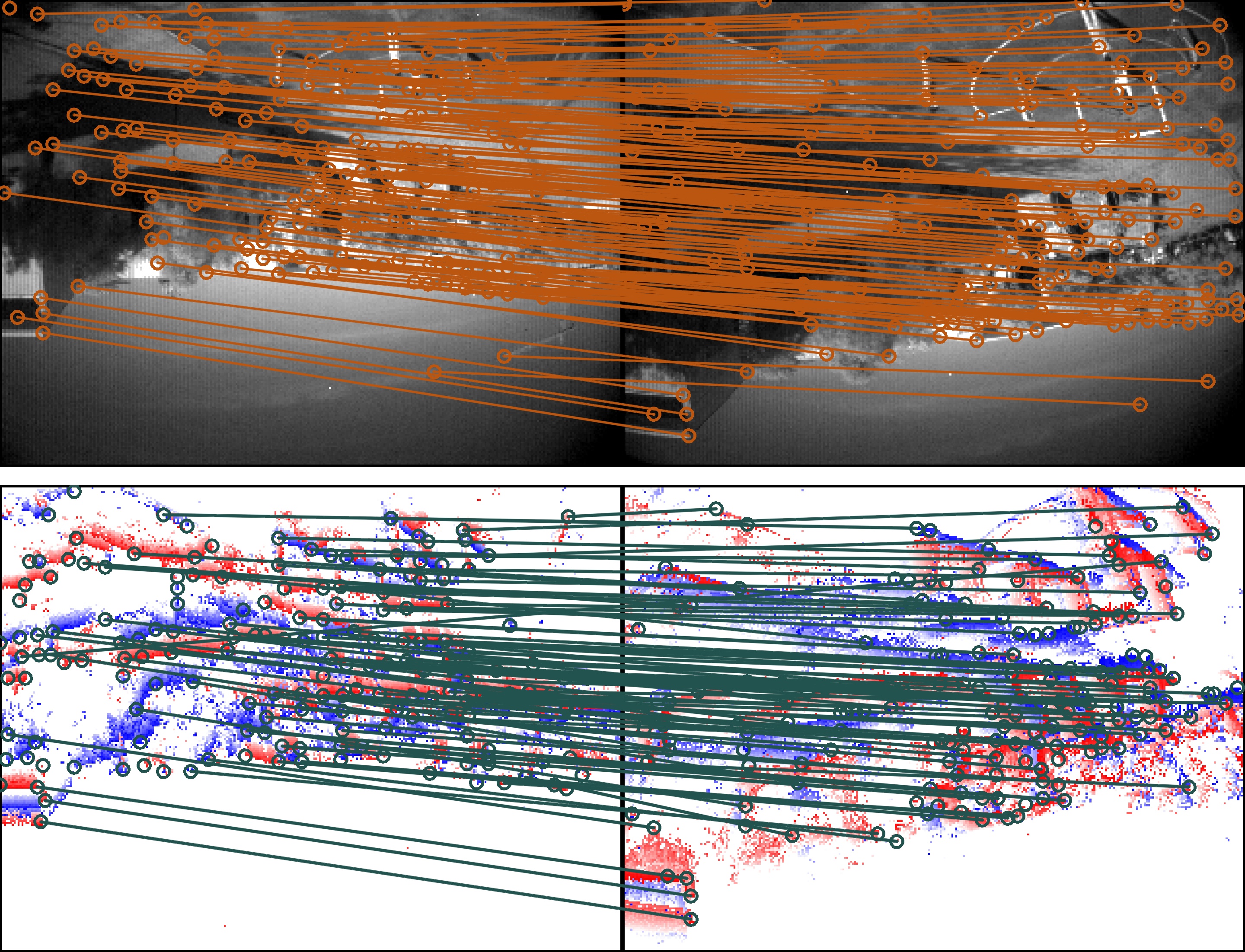}
        \caption{GRIFFIN Perception Dataset~\cite{Rod21}: Soccer People 1}
    \end{subfigure}\vspace{5mm}
    \begin{subfigure}{0.49\textwidth}
        \centering
        \includegraphics[width=0.975\textwidth]{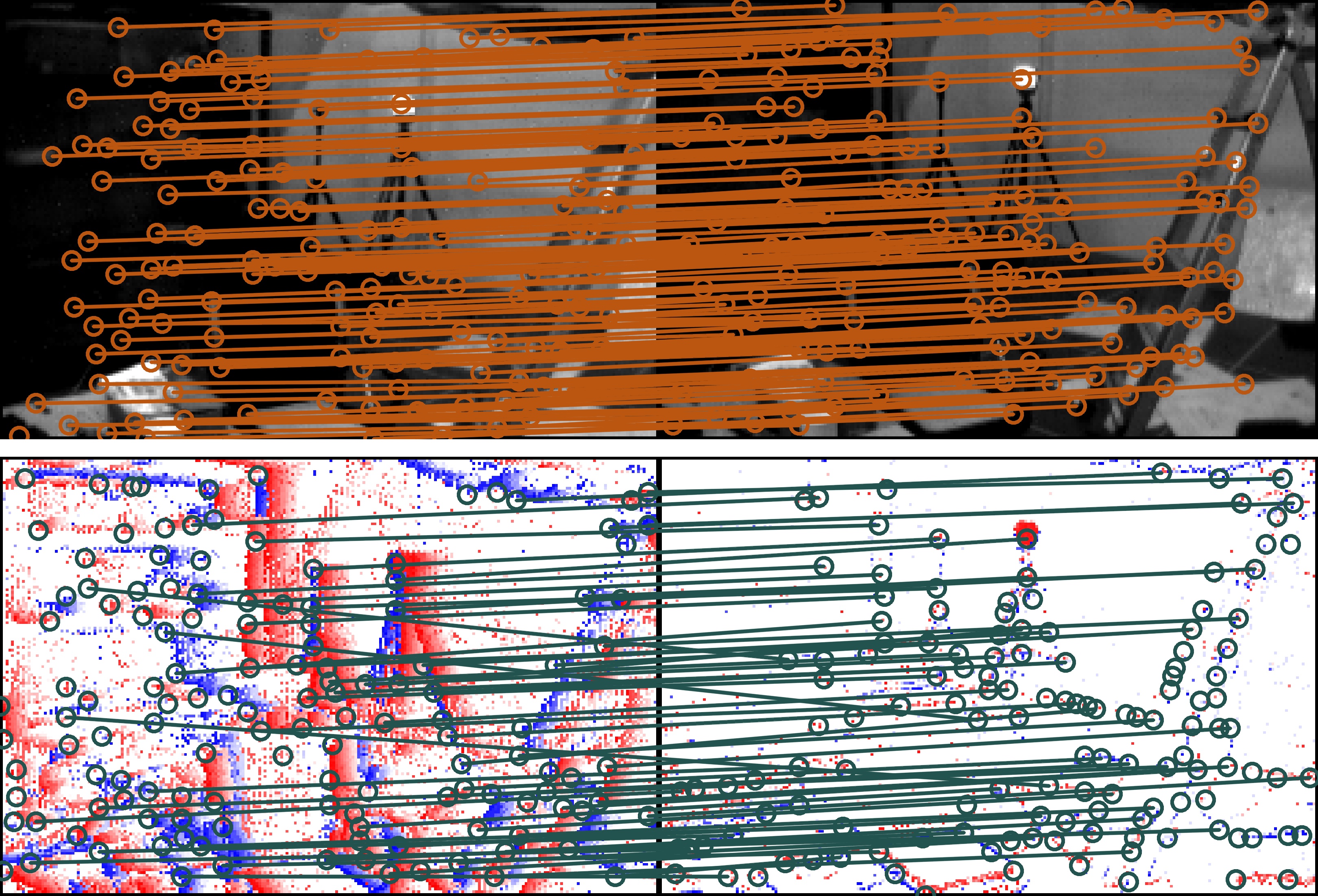}
        \caption{Vision for Visibility Dataset~\cite{Lee22}: Indoor Global Aggressive}
    \end{subfigure}
    \caption{Examples of the training data for temporal matching. Top (orange): pseudo-labels generated by SuperPoint~\cite{Det18} + SuperGlue~\cite{Sar20}; bottom (green): predictions of SuperEvent after training.}
    \label{train_examples}
\end{figure*}

Figure~\ref{train_examples} shows training samples of SuperEvent generated by temporal matching of gray-scale frames. Due to the modality change, SuperEvent does not learn to exactly match the pseudo-labels, but partially detects and matches different keypoints, while still yielding similar patterns.

\begin{figure*}[!ht]
    \centering
    \begin{subfigure}{0.49\textwidth}
        \centering
        \includegraphics[width=0.975\textwidth]{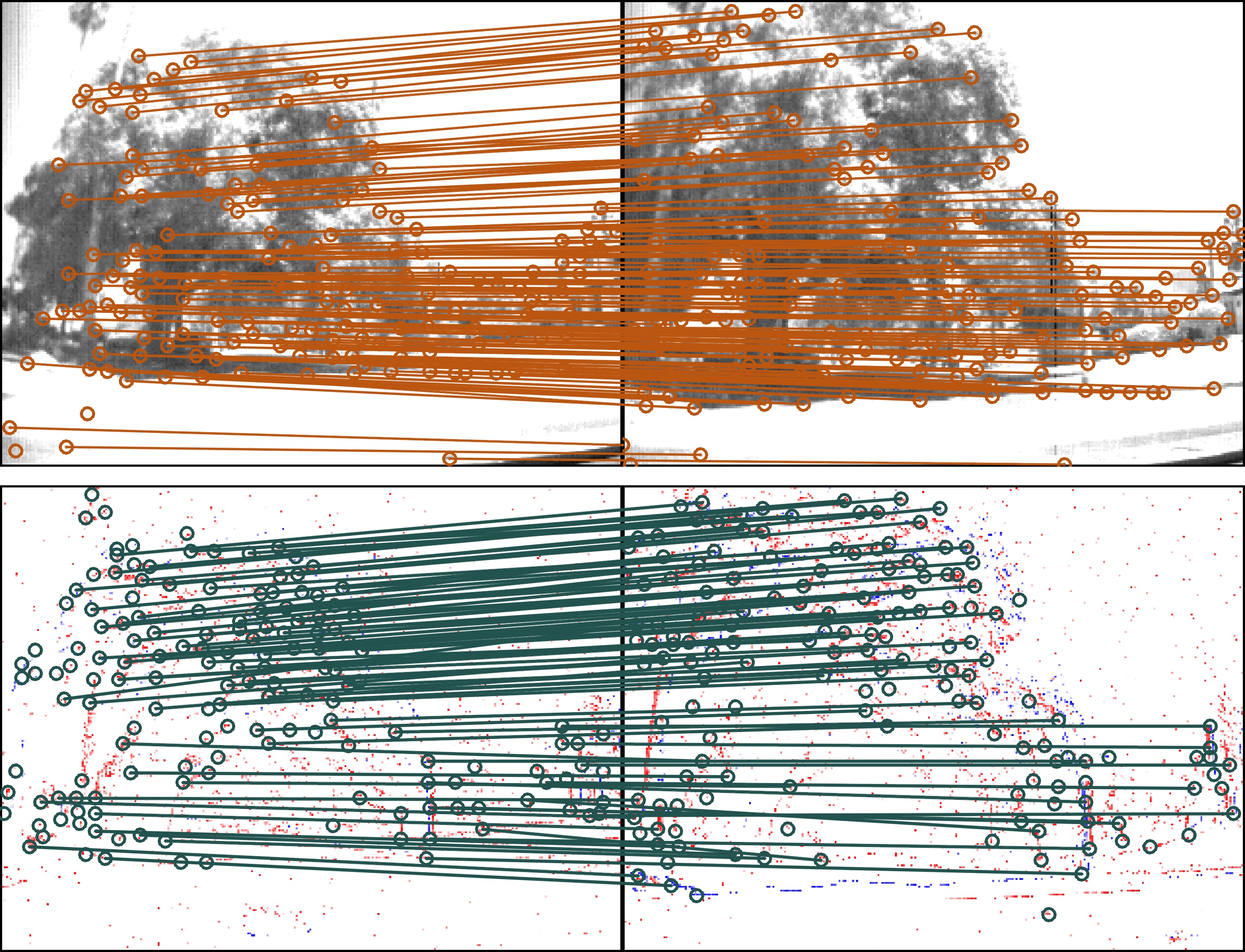}
        \caption{DAVIS Driving Dataset 2020~\cite{Hu20}: rec1501614399}
    \end{subfigure}
    \begin{subfigure}{0.49\textwidth}
        \centering
        \includegraphics[width=0.975\textwidth]{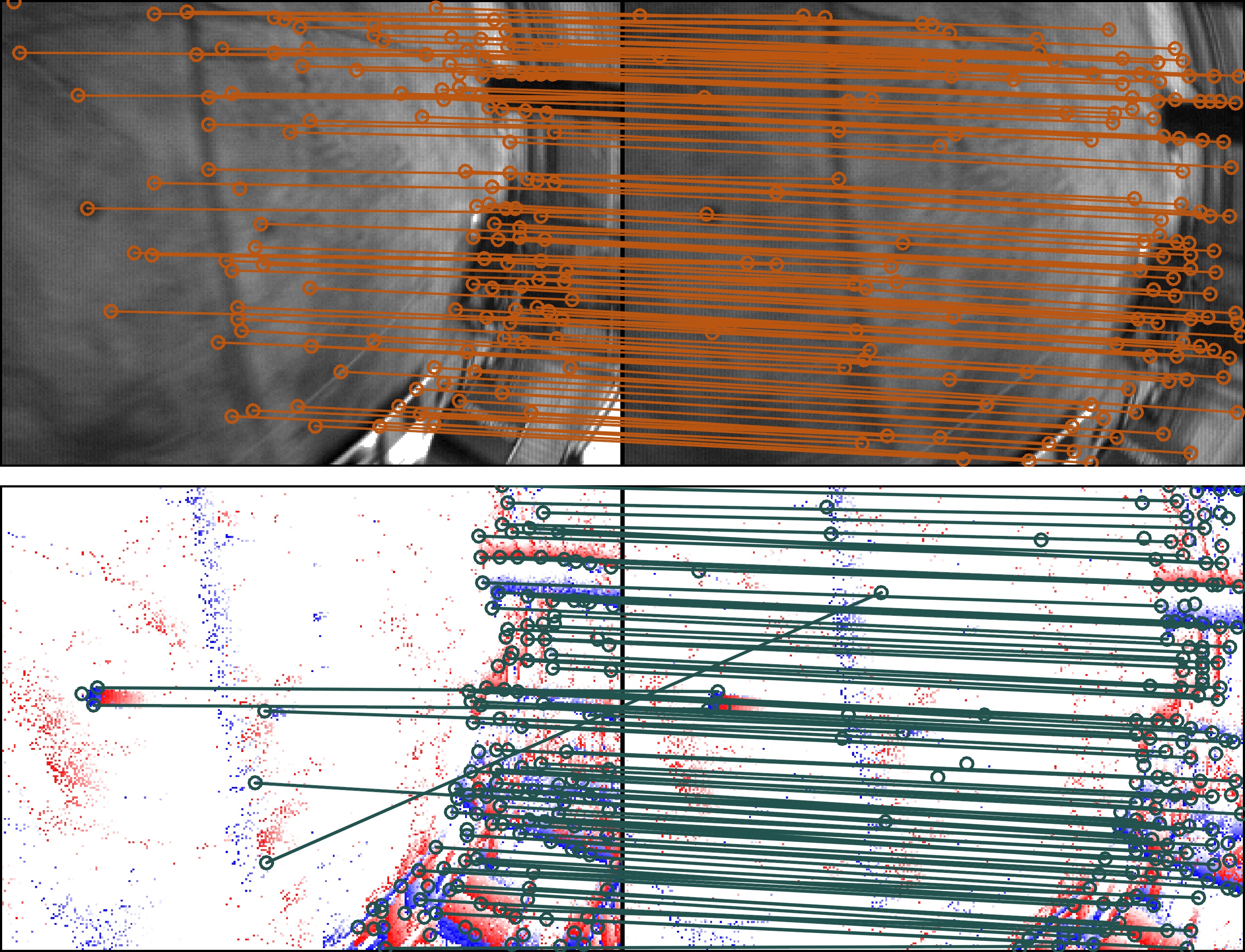}
        \caption{UZH-FPV Drone Racing Dataset~\cite{Del19}: Indoor 45° Downward Facing 14}
    \end{subfigure}\vspace{5mm}
    \begin{subfigure}{0.49\textwidth}
        \centering
        \includegraphics[width=0.975\textwidth]{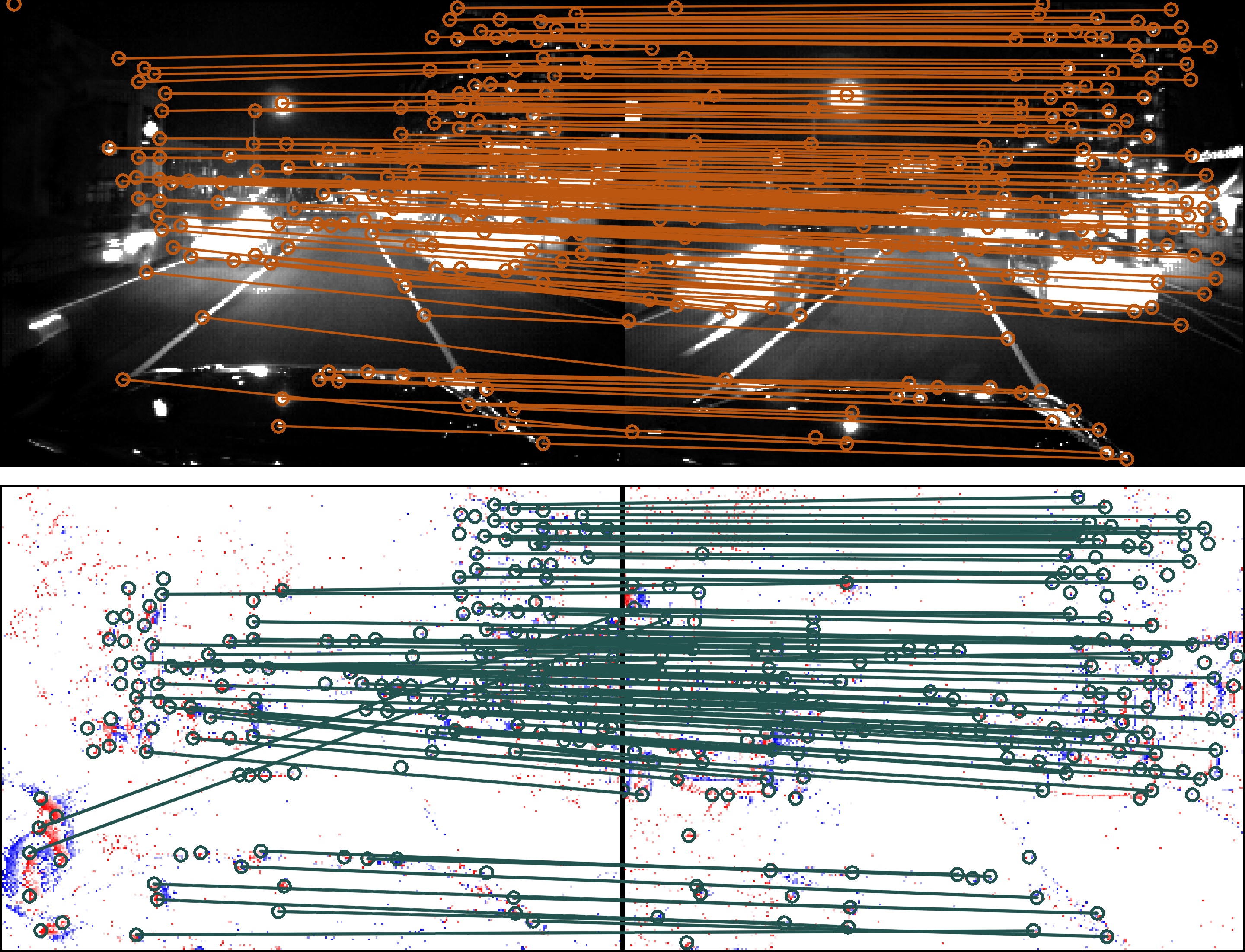}
        \caption{Multi Vehicle Stereo Event Camera Dataset~\cite{Zhu18}: Outdoor Night Drive 1}
    \end{subfigure}
    \begin{subfigure}{0.49\textwidth}
        \centering
        \includegraphics[width=0.975\textwidth]{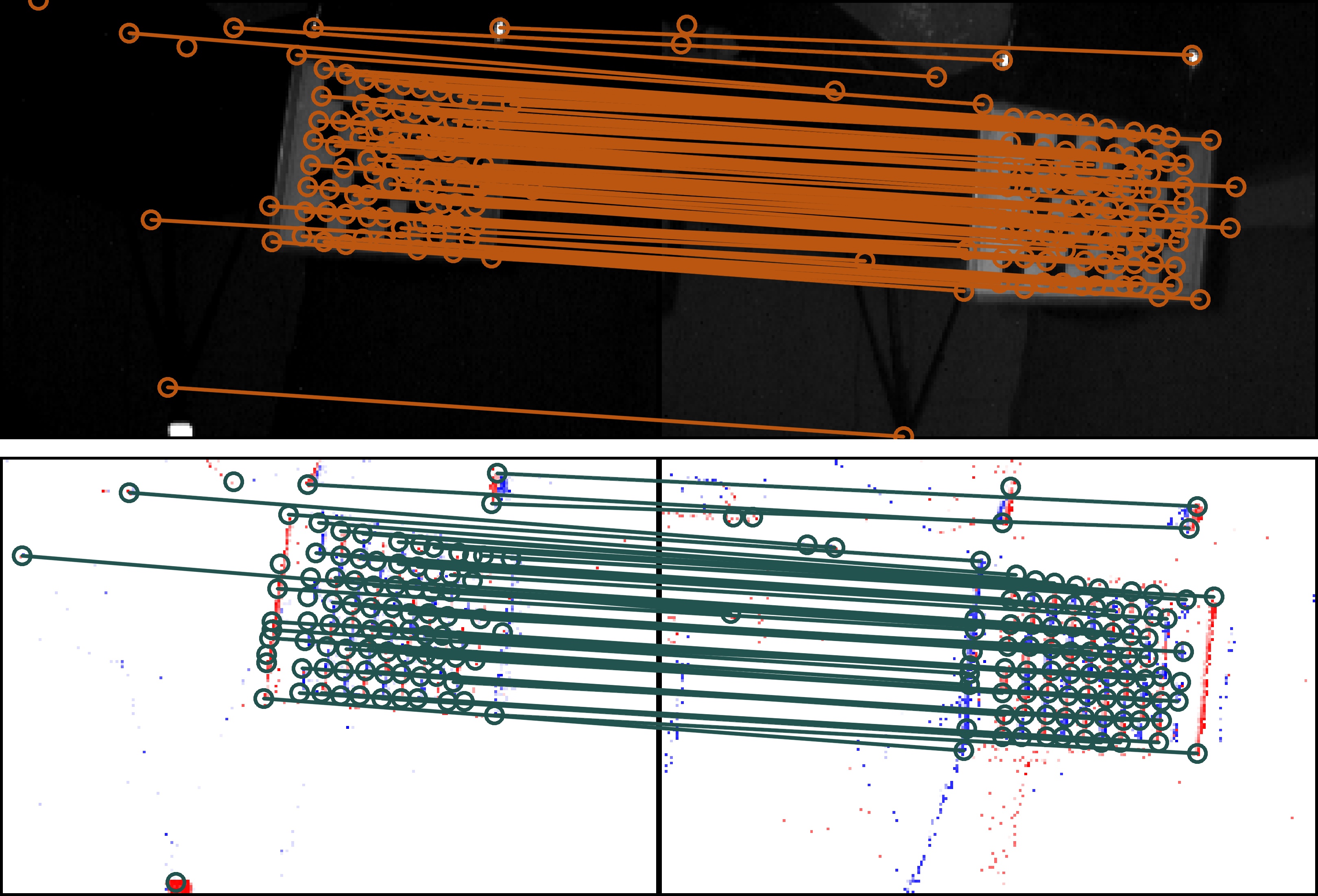}
        \caption{Vision for Visibility Dataset~\cite{Lee22}: Indoor Varying Robust}
    \end{subfigure}
    \caption{Examples of predictions and pseudo-labels of data not used for training. Top (orange): pseudo-labels generated by SuperPoint~\cite{Det18} + SuperGlue~\cite{Sar20}; bottom (green): predictions of SuperEvent.}
    \label{test_examples}
\end{figure*}

Matched keypoints from SuperEvent on unseen sequences are shown in Figure~\ref{test_examples}. These sequences are held out during training, showing the generalization ability of SuperEvent.
\newpage
\section{Event-based versus Frame-based Keypoint Matching}
Since SuperEvent is not explicitly trained in scenarios where the quality of frame cameras degrades, such as fast scene motion and high dynamic range (HDR), we demonstrate its generalization ability by comparing SuperEvent's event-based keypoint correspondences to related frame-based results.

\subsection{Qualitative Comparison for Fast Motion and HDR}
\begin{figure*}[!ht]
    \centering
    \begin{subfigure}{0.49\textwidth}
        \centering
        \includegraphics[width=0.975\textwidth]{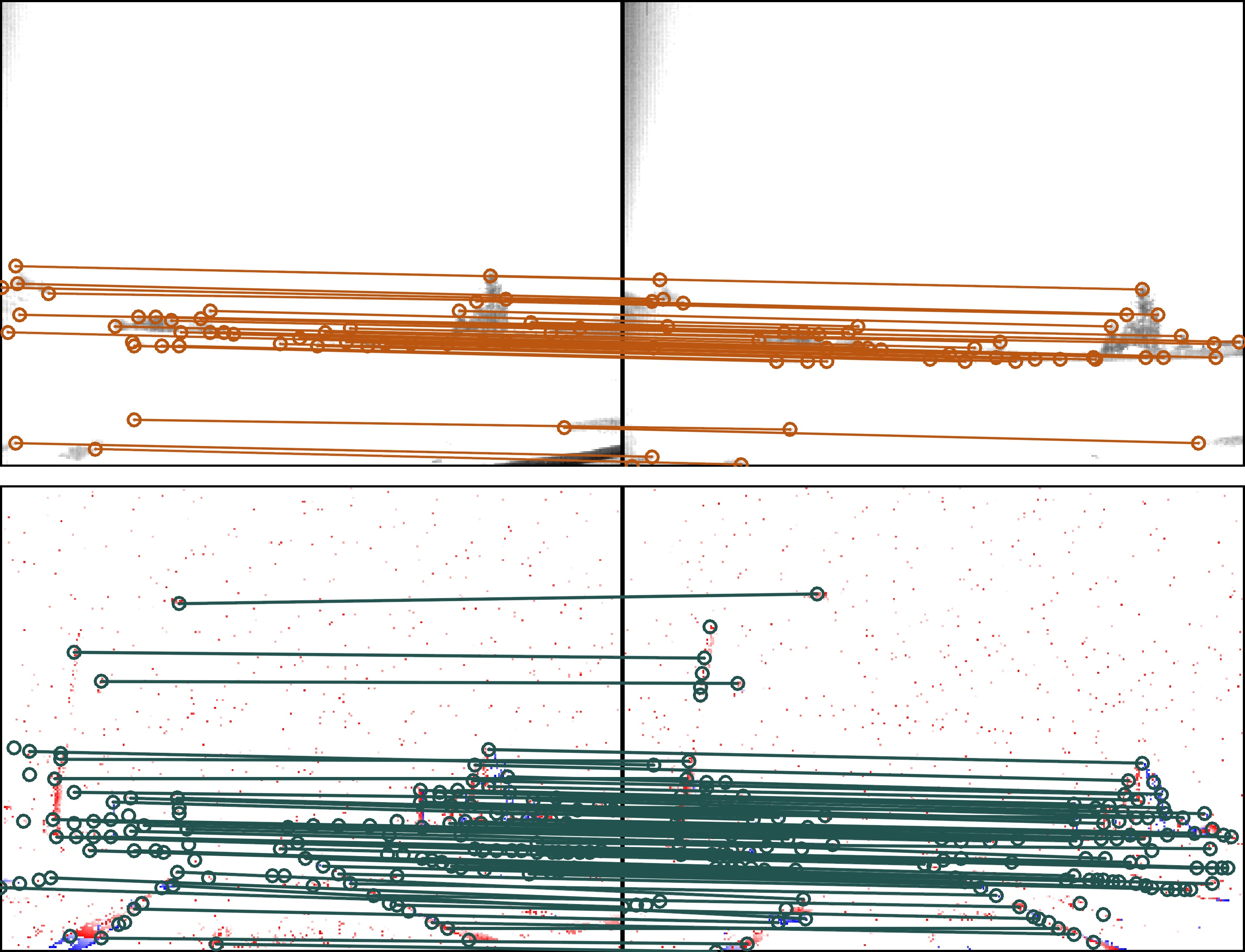}
        \caption{HDR: DAVIS Driving Dataset 2020~\cite{Hu20}: rec1501614399}
    \end{subfigure}\vspace{5mm}
    \begin{subfigure}{0.49\textwidth}
        \centering
        \includegraphics[width=0.975\textwidth]{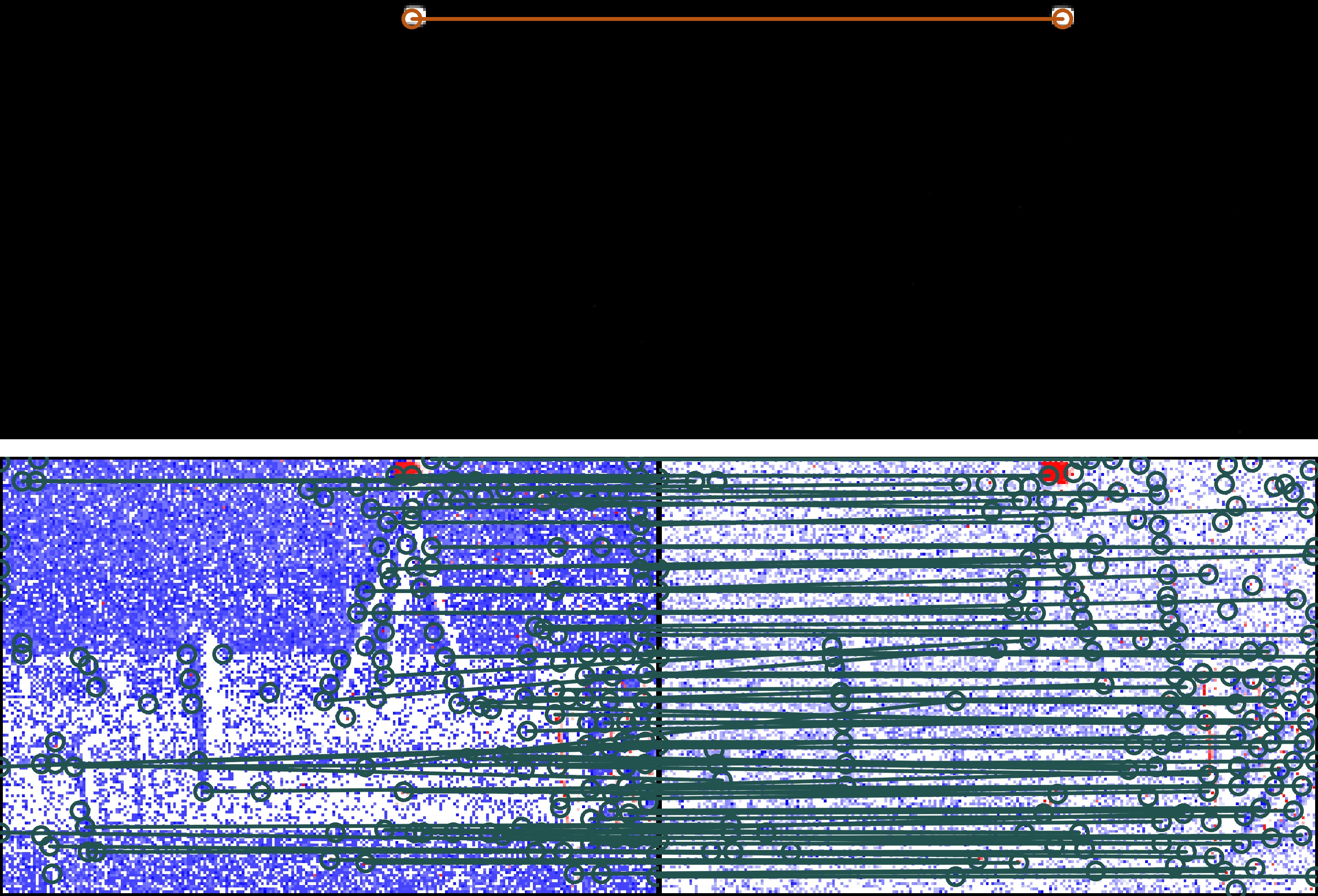}
        \caption{HDR: Vision for Visibility Dataset~\cite{Lee22}: Indoor Varying Robust}
    \end{subfigure}\vspace{5mm}
    \begin{subfigure}{0.49\textwidth}
        \centering
        \includegraphics[width=0.975\textwidth]{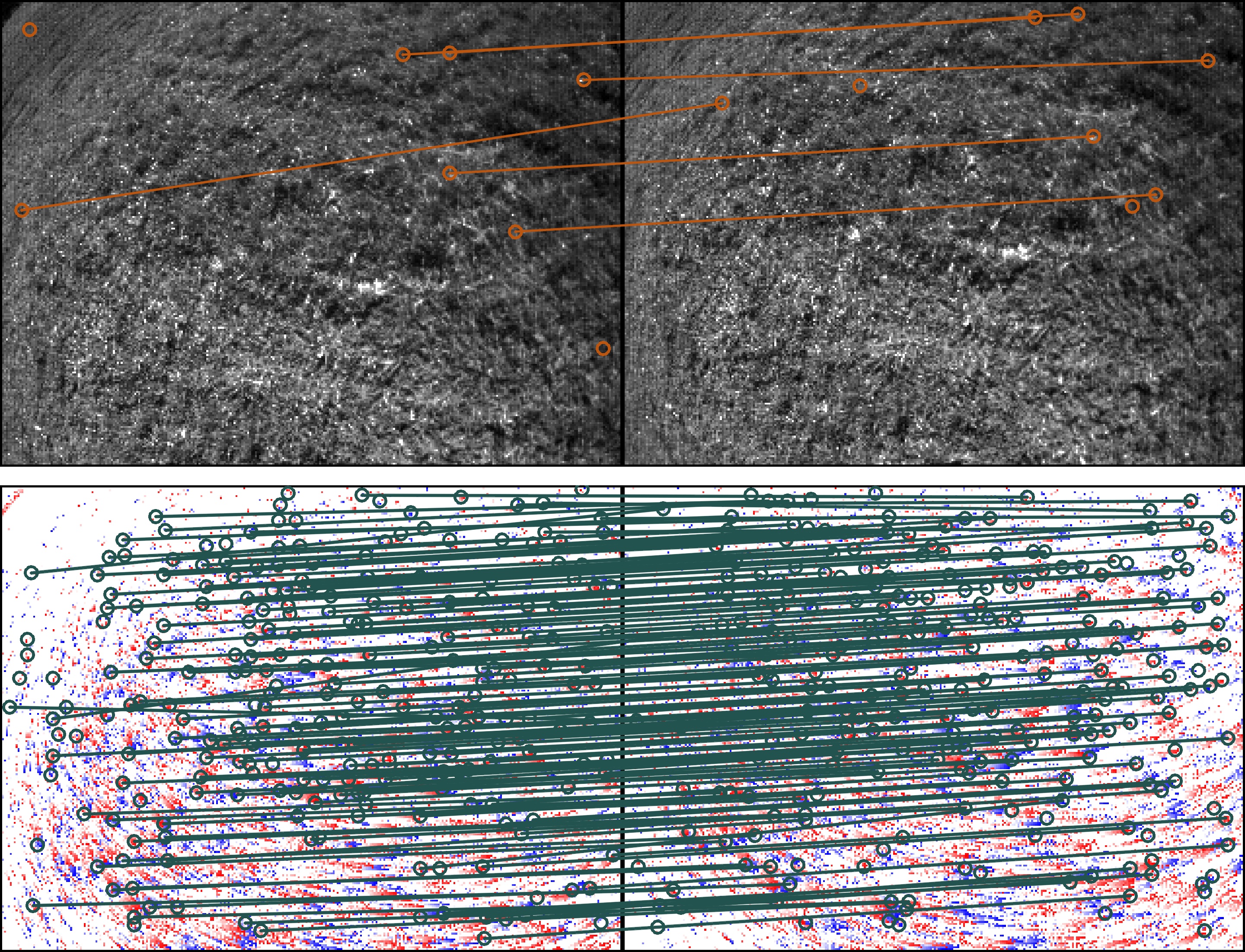}
        \caption{Fast motion: UZH-FPV Drone Racing Dataset~\cite{Del19}: Outdoor 45° Downward Facing 1}
    \end{subfigure}
    \begin{subfigure}{0.49\textwidth}
        \centering
        \includegraphics[width=0.975\textwidth]{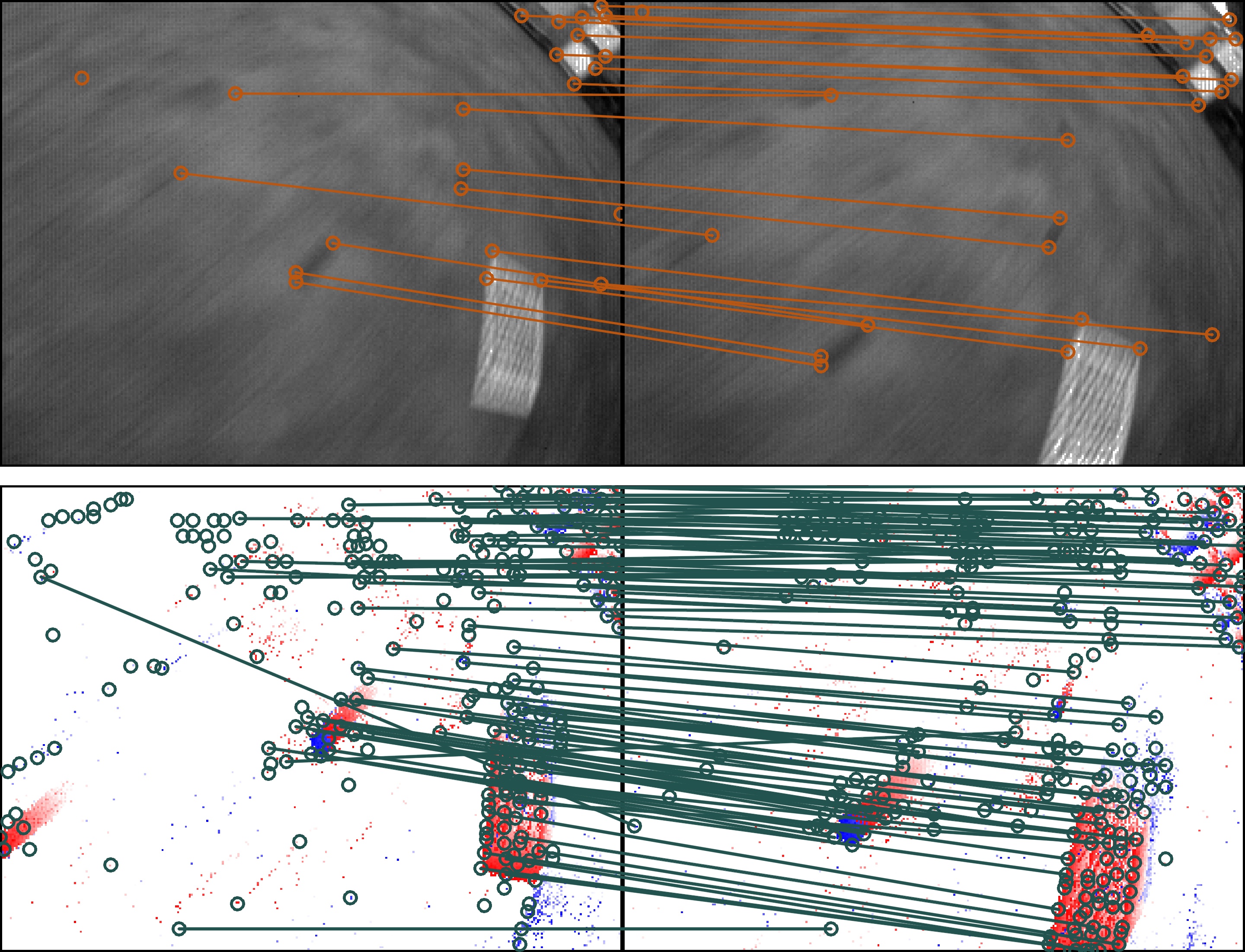}
        \caption{Fast motion: UZH-FPV Drone Racing Dataset~\cite{Del19}: Indoor 45° Downward Facing 14}
    \end{subfigure}
    \caption{Frame-based SuperPoint~\cite{Det18} + SuperGlue~\cite{Sar20} (top), event-based SuperEvent (ours, bottom) on unseen sequences, showing the superior matching capabilities of SuperEvent by leveraging the higher quality of event data for scenes with HDR or fast scene motion.\label{fig:comp}}
\end{figure*}

Figure~\ref{fig:comp} shows matches of the same scenes (not used for training) generated by SuperPoint~\cite{Det18} + SuperGlue~\cite{Sar20} on the frames and by SuperEvent and brute-force matching on the event stream. While the quality of the frames under fast scene motion and HDR degrades, the event stream suffers less under these conditions, resulting in better and more equally distributed keypoint matches when using SuperEvent.

\subsection{Comparison in SLAM Downstream Task}
We quantitatively compare the SLAM results from plain frame-based OKVIS2~\cite{Leu22} and our modified version after replacing the frame-based BRISK~\cite{Leu11} detector with SuperEvent. Since event data degrades less than frames in such conditions, SuperEvent's predictions improve OKVIS2's estimations. Choosing a higher processing rate further boosts the performance: the visual overlap between MCTSs rises, and the effects of motion dependence are mitigated.

\begin{table*}[t]
\begin{center}
\vspace{5mm}
\caption{Results on TUM-VIE~\cite{Kle21} sequences with fast motion (\textit{mocap-shake}) and low light (\textit{floor2-dark}). ATE and RPE in cm; RPE for consecutive frames. Results marked $^*$ are not representative due to discontinuous ground truth.\label{tumvie}}
\begin{tabularx}{\linewidth}{
l
>{\centering}X
>{\centering}X
>{\centering}X
>{\centering}X
>{\centering}X
>{\centering\arraybackslash}X
}
\toprule
OKVIS2~\cite{Leu22} & \multicolumn{4}{c}{\textbf{+ SuperEvent (ours)}} & \multicolumn{2}{c}{frame-based} \\
\cmidrule(lr){1-1}
\cmidrule(lr){2-5}
\cmidrule(lr){6-7}
 & \multicolumn{2}{c}{20 Hz} & \multicolumn{2}{c}{40 Hz} & \multicolumn{2}{c}{20 Hz} \\
\cmidrule(lr){2-3}
\cmidrule(lr){4-5}
\cmidrule(lr){6-7}
Sequence & ATE & RPE & ATE & RPE & ATE & RPE \\
\midrule

mocap-shake & \underline{43.71} & \underline{0.55} & \textbf{29.14} & \textbf{0.26} & 50.83 & 1.03 \\

mocap-shake2 & \underline{43.75} & \underline{0.80} & \textbf{27.37} & \textbf{0.46} & 66.29 & 1.39 \\

floor2-dark & \underline{9.58} & ~~\underline{2.51}$^*$ & \textbf{9.37} & ~~\textbf{1.23}$^*$ & failed & failed \\
\bottomrule
\end{tabularx}
\end{center}
\end{table*}
\section{Pose Estimation Experiment}
In this section, we explain the details of the keypoint-based pose estimation benchmark and justify why we chose this method to evaluate SuperEvent.

\subsection{Benchmark Design}
Pose Estimation requires reliable keypoint detection and matching and is therefore a common baseline for frame-based keypoint detectors~\cite{Sar20, Tys20, Sun21, Gle23}. It also indicates the approaches' usability for downstream applications such as Visual Odometry, Simultaneous Localization and Mapping (SLAM), and Structure-from-Motion (SfM) that usually rely on keypoint-based pose estimation. Frame-based approaches are evaluated on datasets such as ScanNet~\cite{Dai17} and MegaDepth~\cite{Li18} containing various images of the same scene with the associated ground truth camera poses, allowing for a straightforward pose estimation evaluation. However, event datasets are usually temporally continuous sequences because of the sensor's asynchronous nature. This raises the question, at which timestamps the pose estimation is evaluated. We define the evaluation benchmark as follows:
\begin{itemize}
    \item At each time step $t_{i}$ with available ground truth data, the ground truth camera orientation must change by the maximal rotation change angle $c_{\Delta \text{r}, \text{max}}$ within the maximal time difference $c_{\Delta \tau}$.
    \item We find equally $n$ distributed rotation changes in this time interval to evaluate the pose estimation.
\end{itemize}
In this experiment, we choose $c_{{\Delta \text{r}, \text{max}}} = \SI{45}{\text{°}}$ $c_{{\Delta \tau}} = \SI{2}{s}$ and $n_{\Delta \text{r}} = 45$ to test various levels of difficulty while reducing the amount of samples without visual overlap. Our practical implementation executes the following steps for each sequence:
\begin{enumerate}
    \item For all timestamps with associated ground truth pose measurements (usually between 100 and \SI{200}{Hz}), we generate keypoint and descriptor predictions. For tracking approaches, we assign a track ID as a scalar descriptor. Matching the same track ID reproduces the tracking result.
    \item Next, we iteratively check for each ground truth sample if any of the subsequent ground truth samples within $c_{{\Delta \tau}} = \SI{2}{s}$ yields an rotation change of at least $c_{{\Delta \text{r}, \text{max}}} = \SI{45}{\text{°}} \}$. If this condition cannot be fulfilled, we skip the respective sample.
    \item For samples with sufficient rotation change within $c_{{\Delta \text{r}, \text{max}}}$, we find the first subsequent ground truth rotation values that surpass the equally distributed rotation changes $\frac{c_{\Delta \text{r}, \text{max}}}{n_{\Delta \text{r}}} \cdot \{1, 2, \ldots, n_{\Delta \text{r}}\} = \{ \SI{1}{\text{°}}, \SI{2}{\text{°}}, \ldots, \SI{45}{\text{°}} \}$.
    \item We match the keypoint descriptors for these selected prediction samples with associated ground truth measurements. The camera pose is estimated based on the resulting keypoint pairs and considering the lens distortion (unless the approach to evaluate already required a rectified event representation as input). We calculate the rotation difference and its angle of the axis-angle representation. 
    \item Having evaluated samples of the dataset sequences, we report the area-under-curve (AUC) for different thresholds as in~\cite{Sun21}.
\end{enumerate}
We evaluate SuperEvent on the following sequences, omitting the ones without sufficient rotation changes as well as calibration sequences.\\
Event Camera Dataset~\cite{Mue17_ecd}: 
\begin{itemize}
\item boxes\_6dof
\item boxes\_rotation
\item poster\_6dof
\item poster\_rotation
\item shapes\_6dof
\item shapes\_rotation
\end{itemize}
Event-aided Direct Sparse Odometry~\cite{Hid22}:
\begin{itemize}
\item peanuts\_dark
\item peanuts\_light
\item rocket\_earth\_light
\item rocket\_earth\_dark
\item ziggy\_and\_fuzz
\item ziggy\_and\_fuzz\_hdr
\item peanuts\_running
\item all\_characters
\end{itemize}

\subsection{Why Not Homography Estimation?}

\begin{figure}[b]
    \centering
    \includegraphics[width=0.8\columnwidth]{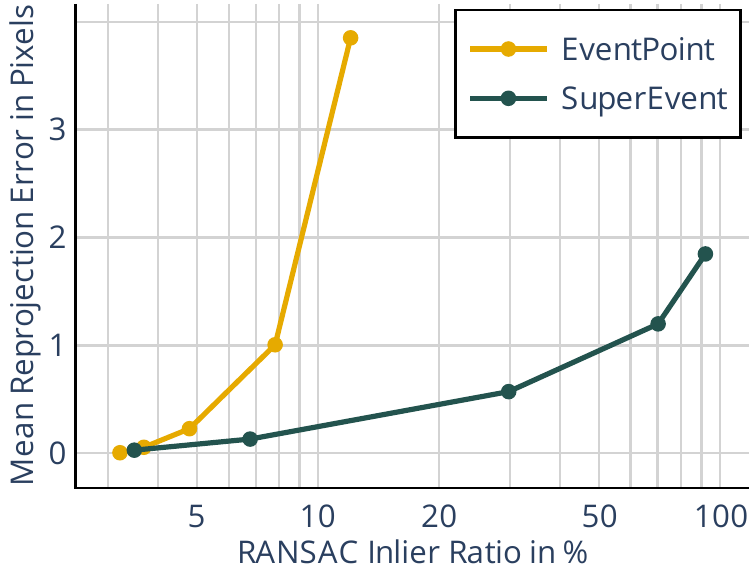}
    \caption{Homography estimation on HVGA ATIS~\cite{Man19} matching 400 keypoints per sample, $\Delta t = \SI{50}{ms}$, and \textit{RANSAC reprojection error thresholds} of \{0.1, 0.3, 1, 3, 10\} pixels.\label{fig:hom}}
\end{figure}

Some of the existing approaches are benchmarked on homography estimation of planar scenes~\cite{Man19, Chi21, Chi22, Hua23, Gao24}. However, the commonly used HVGA ATIS Corner dataset~\cite{Man19} contains neither ground truth poses nor homography measurements. Therefore, the authors compare the mean reprojection error (MRE) of their method's detected points warped by the estimated homography. But without comparing it to any ground truth, in general, it cannot be guaranteed that the estimated homography is (close to) correct. E.g., in most cases, a nonsensical estimate that matches four random keypoints can achieve the optimal score of 0 since this is the minimum number of point correspondences to estimate the homography; and there will not be any outliers that negatively influence the score.\par
As long as there are always sufficiently many keypoints detected, this evaluation procedure might still be sensible for approaches that rely on basic nearest neighbor matching in pixel-space because some wrong matches do not lead to large errors. However, for approaches that rely on descriptor matching~\cite{Man19, Chi21, Chi22}, only a few outliers with large errors have a serious negative impact on the reported score. Most downstream applications therefore employ outlier filtering, such as Random Sample Consensus (RANSAC) ~\cite{Fis81}.  Also, the approaches relying on descriptor matching~\cite{Hua23, Gao24} employ RANSAC as their homography estimation benchmark. The RANSAC algorithm rejects outliers with a reprojection error greater than a pre-defined threshold $c_{\text{R}}$. Thus, this threshold is an upper bound to the estimated homography reprojection error since all keypoints with larger errors are filtered out. Thereby, the MRE score can be arbitrarily reduced by choosing a smaller $c_{\text{RE}}$, making it inappropriate for performance benchmarking.\par
This general problem applies not only to RANSAC but to all approaches relying on some form of outlier removal as a post-processing step. Outliers, of course, are never completely avoidable, and it is a common procedure to remove them in downstream applications. Therefore, we decided to reproduce the frame-based benchmark of estimating the camera pose change in datasets with ground truth camera pose measurements -- producing meaningful results for approaches with outlier removal.\par
We illustrate this issue in Figure~\ref{fig:hom} where we plot the reprojection error after homography estimation for different \textit{RANSAC reprojection error thresholds} over the ratio of matches classified as inliers: SuperEvent achieves similar reprojection errors as EventPoint~\cite{Hua23} with RANSAC leading to far fewer outliers.
\section{Stereo Event-Visual Intertial SLAM Experiment}

\begin{table}[t]
\begin{center}
\caption{RPE scores in cm for consecutive samples at 20 Hz of OKVIS2~\cite{Leu22} + SuperEvent on the TUM-VIE \textit{mocap} sequences.\label{tab:tum_rpe}}
\small
\begin{tabularx}{\linewidth}{
>{\centering}X
>{\centering}X
>{\centering}X
>{\centering}X
>{\centering\arraybackslash}X
}
\toprule
1d-trans & 3d-trans & 3dof & desk & desk2 \\
\midrule
0.07 & 0.14 & 0.10 & 1.39 & 1.25 \\
\bottomrule
\end{tabularx}
\vspace{-12pt}
\end{center}
\end{table}

Lastly, we visualize 2D projections of the trajectories estimated by SuperEvent integrated into OKVIS2~\cite{Leu22} yielding the reported results. In addition to the ATE reported in the main paper, we also report the RPE results of our method for completeness.

\subsection{TUM-VIE Small-scale Sequences}
\begin{figure*}[!hbt]
    \centering
    \vspace{20mm}
    \begin{subfigure}{0.24\textwidth}
        \centering
        \includegraphics[height=0.34\textheight]{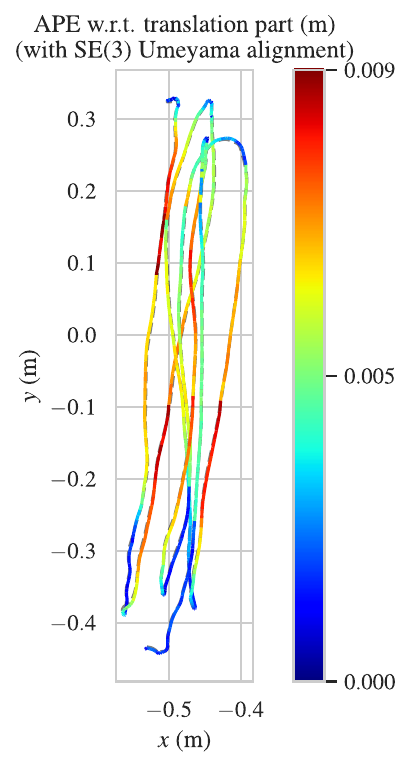}
        \caption{mocap\_1d-trans\vspace{5mm}}
    \end{subfigure}
    \begin{subfigure}{0.29\textwidth}
        \centering
        \includegraphics[height=0.34\textheight]{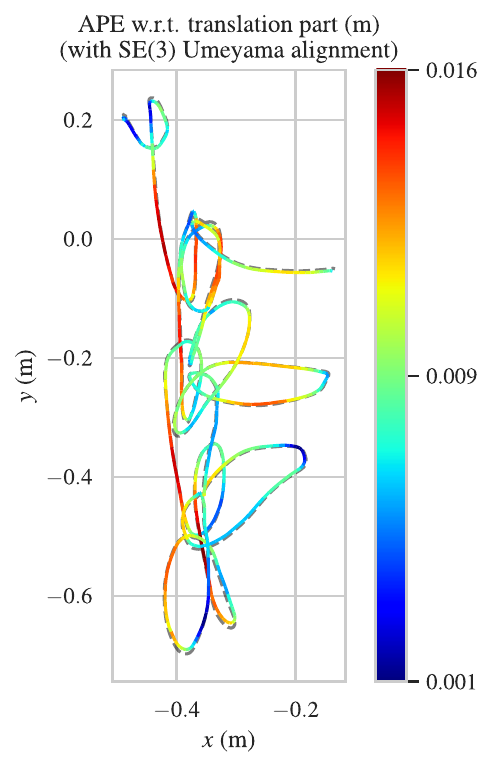}
        \caption{mocap\_3d-trans\vspace{5mm}}
    \end{subfigure}
    \begin{subfigure}{0.46\textwidth}
        \centering
        \includegraphics[height=0.34\textheight]{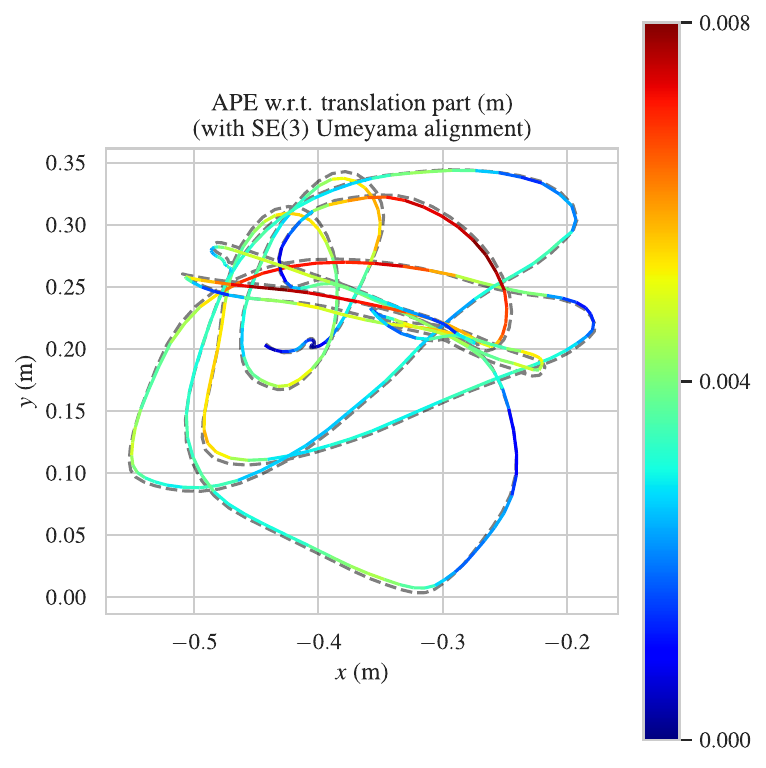}
        \caption{mocap\_6dof\vspace{5mm}}
    \end{subfigure}
    \begin{subfigure}{0.49\textwidth}
        \centering
        \includegraphics[height=0.34\textheight]{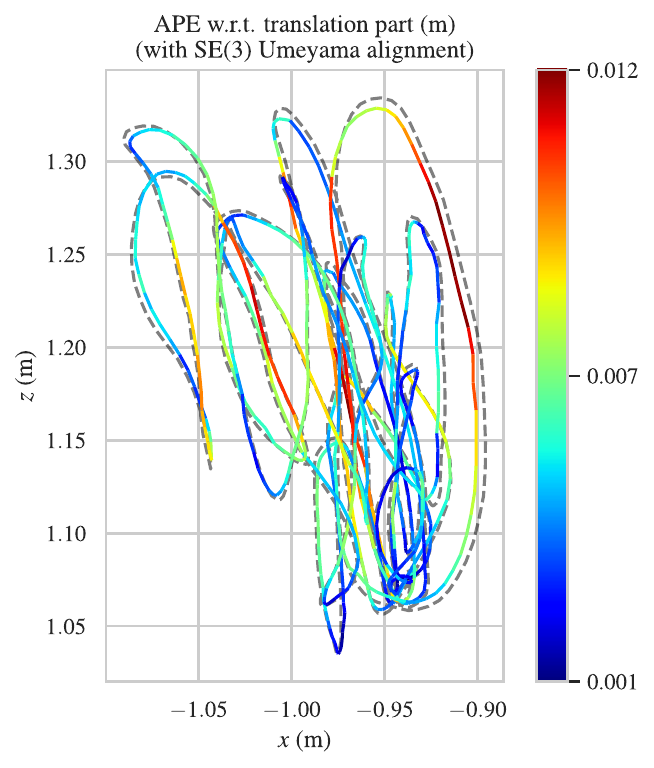}
        \caption{mocap\_desk}
    \end{subfigure}
    \begin{subfigure}{0.49\textwidth}
        \centering
        \includegraphics[height=0.34\textheight]{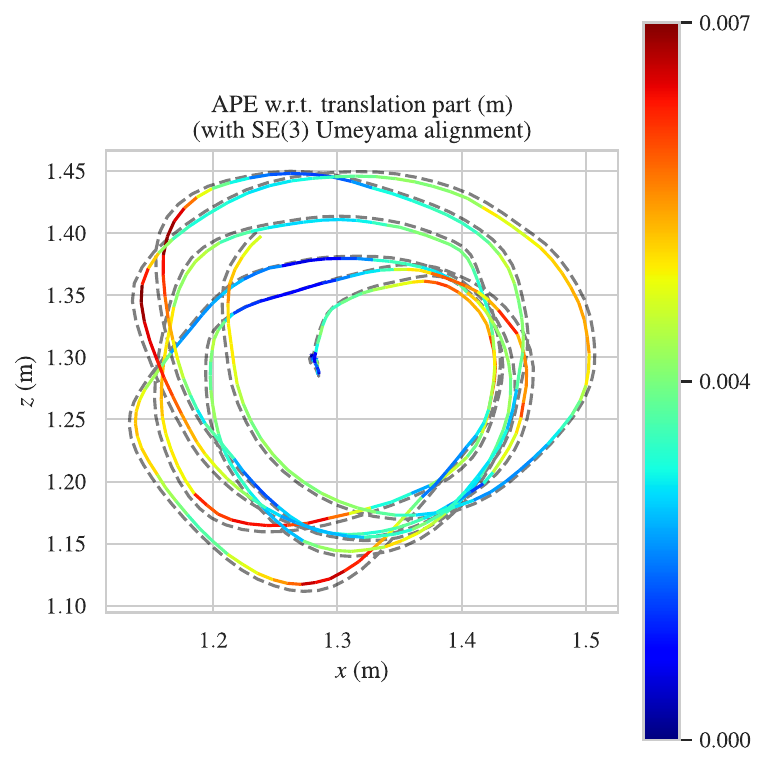}
        \caption{mocap\_desk2}
    \end{subfigure}
    \caption{OKSVIS2 + SuperEvent's trajectories on the TUM-VIE \textit{mocap} sequences with ground truth trajectory (dashed) and absolute position error of the 3D trajectories shown by the color.}
    \label{tumvie_small}
    \vspace{20mm}
\end{figure*}

Figure~\ref{tumvie_small} shows the trajectories from SuperEvent + OKVIS2 on the TUM-VIE~\cite{Kle21} \textit{mocap}-sequences. Since OKVIS2 is non-deterministic, we process every sequence 5 times and select the trajectory with median error. We report the RPE scores in Table~\ref{tab:tum_rpe}.\par

\subsection{TUM-VIE Large-scale Sequences}
\begin{figure*}[!hbt]
    \centering
    \begin{subfigure}{0.75\textwidth}
        \includegraphics[width=\textwidth]{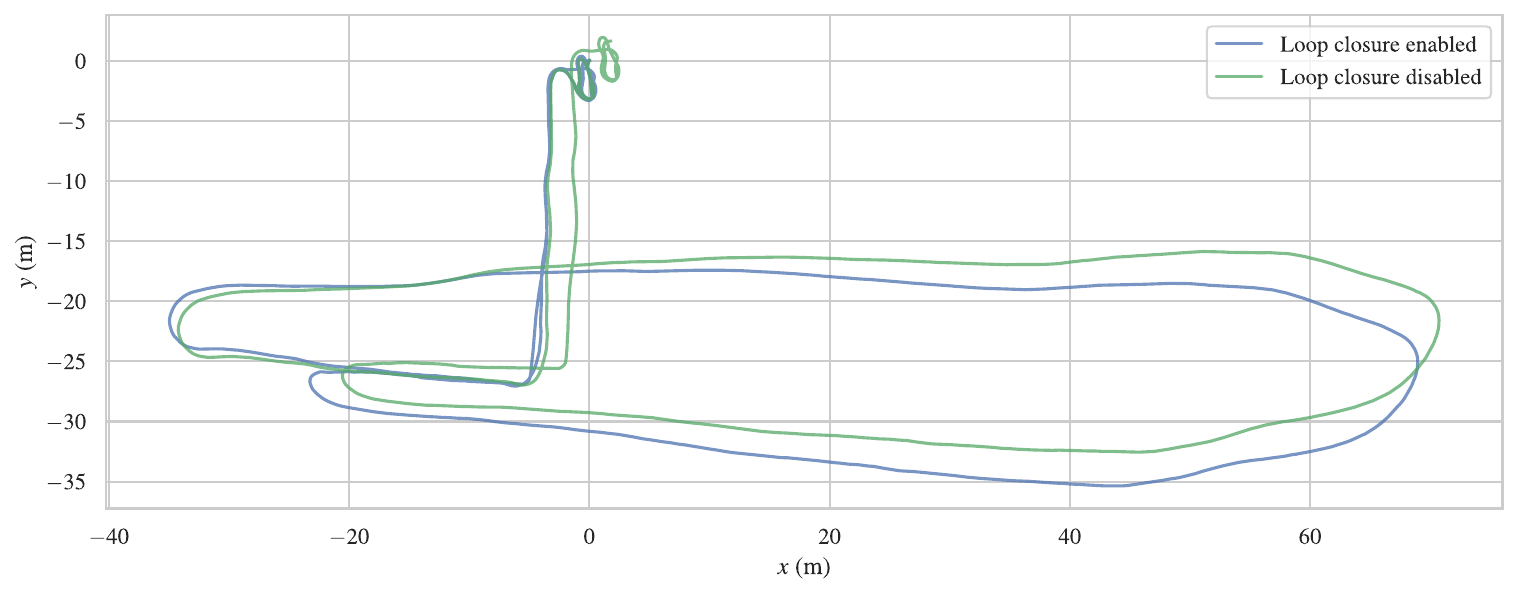}
    \end{subfigure}
    \begin{subfigure}{0.75\textwidth}
        \includegraphics[width=\textwidth]{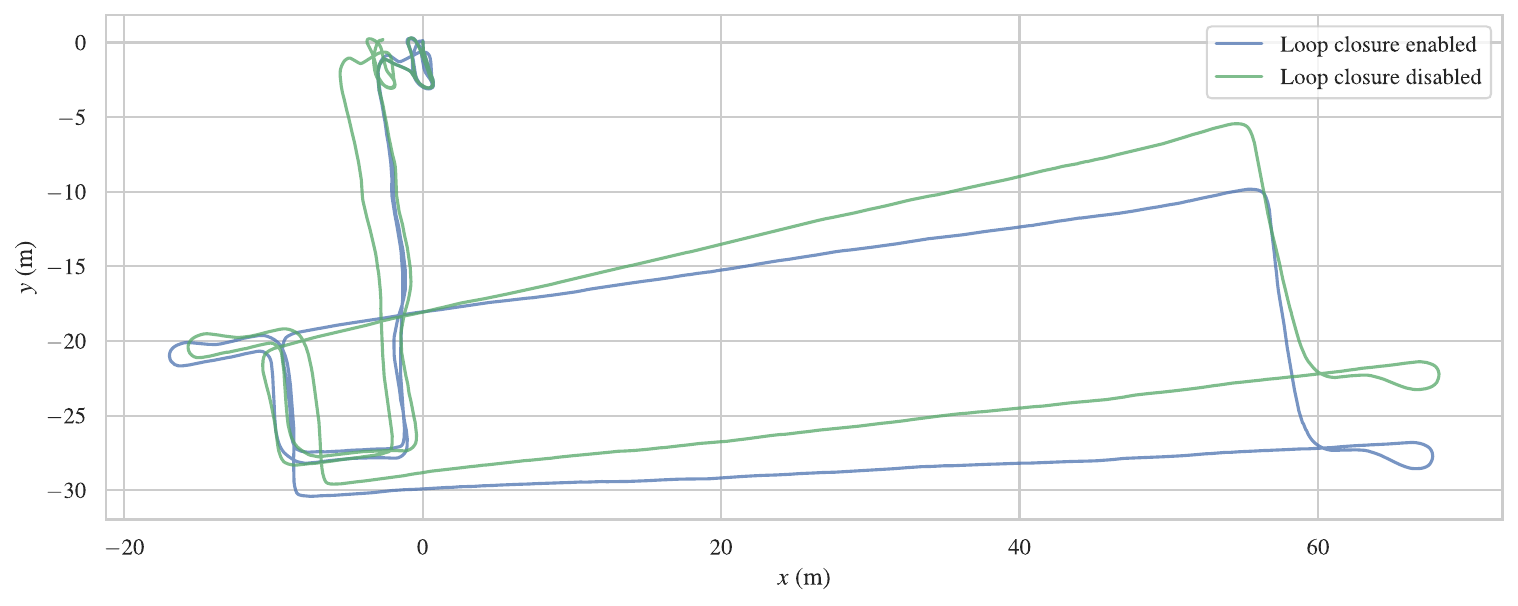}
    \end{subfigure}
    \begin{subfigure}{0.75\textwidth}
        \includegraphics[width=\textwidth]{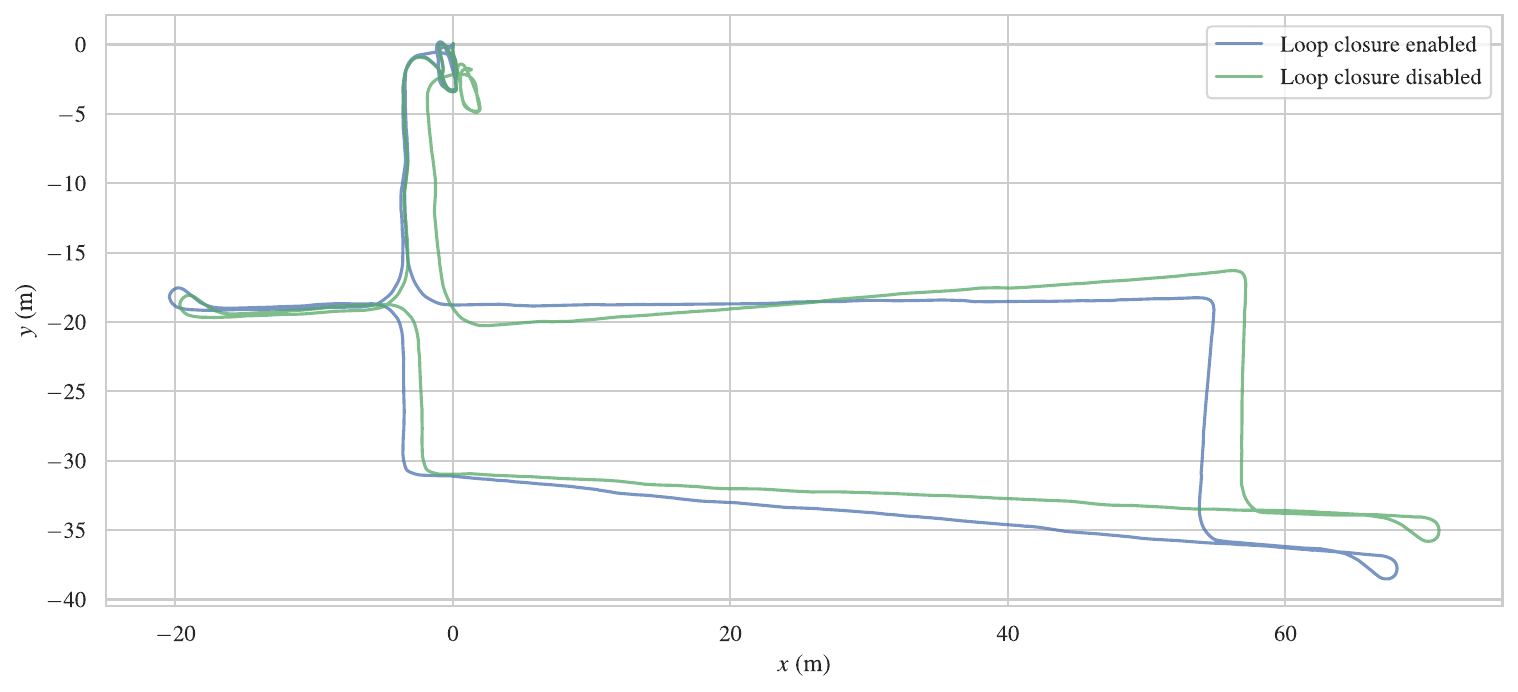}
    \end{subfigure}
    \begin{subfigure}{0.75\textwidth}
        \includegraphics[width=\textwidth]{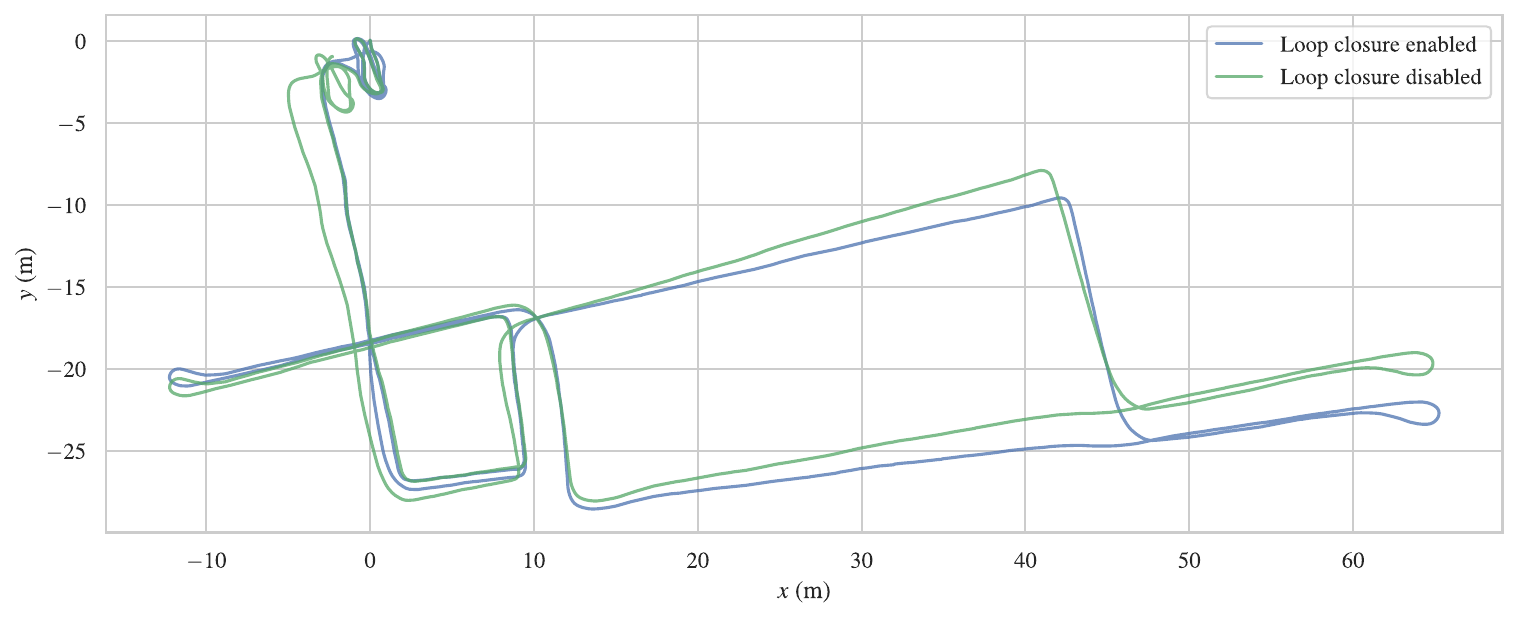}
    \end{subfigure}
    \caption{OKSVIS2 + SuperEvent's trajectories on the TUM-VIE loop-floor 0-3 sequences with and without loop closure.}
    \label{tumvie_large}
\end{figure*}

The effect of loop closure on the trajectory estimation of OKVIS2 + SuperEvent can be seen in Figure~\ref{tumvie_large}. The loop closure is reliably detected on all 4 \textit{loop-floor} sequences of the TUM-VIE dataset. Since the ground truth is not continuous, we do not report RPE scores for these sequences, as they lack interpretive value and are not comparable.\par

\subsection{VECtor Large-scale Sequences}
\begin{figure*}[!hbt]
    \centering
    \vspace{25mm}
    \begin{subfigure}{0.405\textwidth}
        \centering
        \includegraphics[height=0.3125\textheight]{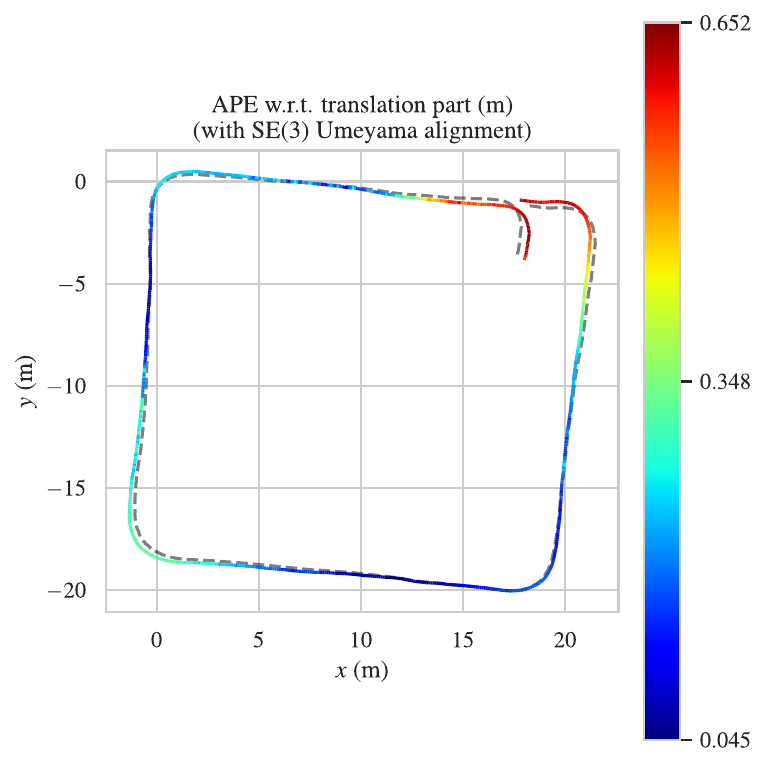}
        \caption{corridors-dolly\vspace{5mm}}
    \end{subfigure}
    \begin{subfigure}{0.295\textwidth}
        \centering
        \includegraphics[height=0.3125\textheight]{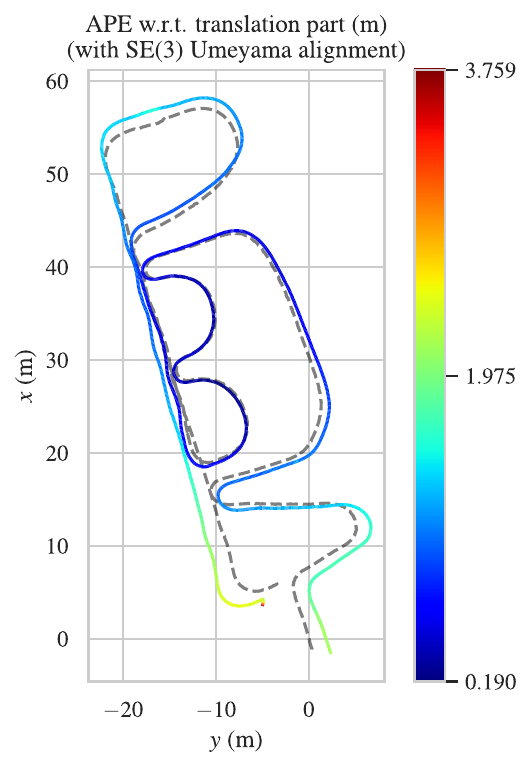}
        \caption{units-dolly\vspace{5mm}}
    \end{subfigure}
    \begin{subfigure}{0.29\textwidth}
        \centering
        \includegraphics[height=0.3125\textheight]{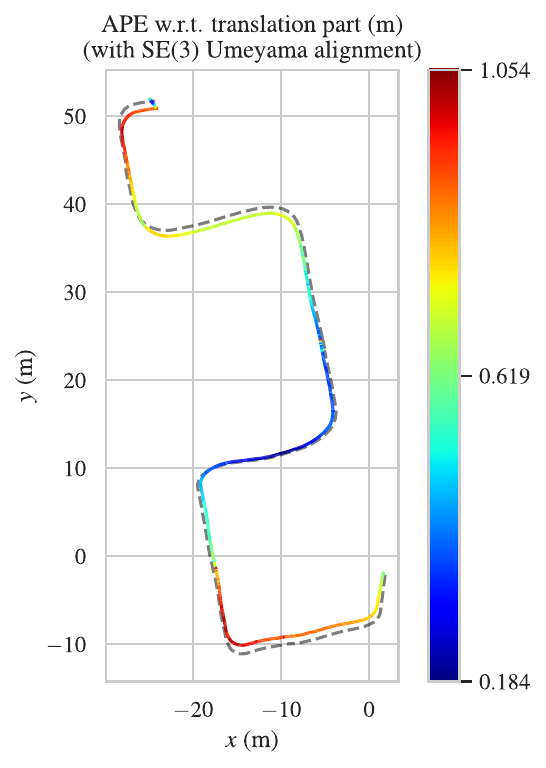}
        \caption{school-dolly\vspace{5mm}}
    \end{subfigure}
    \begin{subfigure}{0.405\textwidth}
        \centering
        \includegraphics[height=0.3125\textheight]{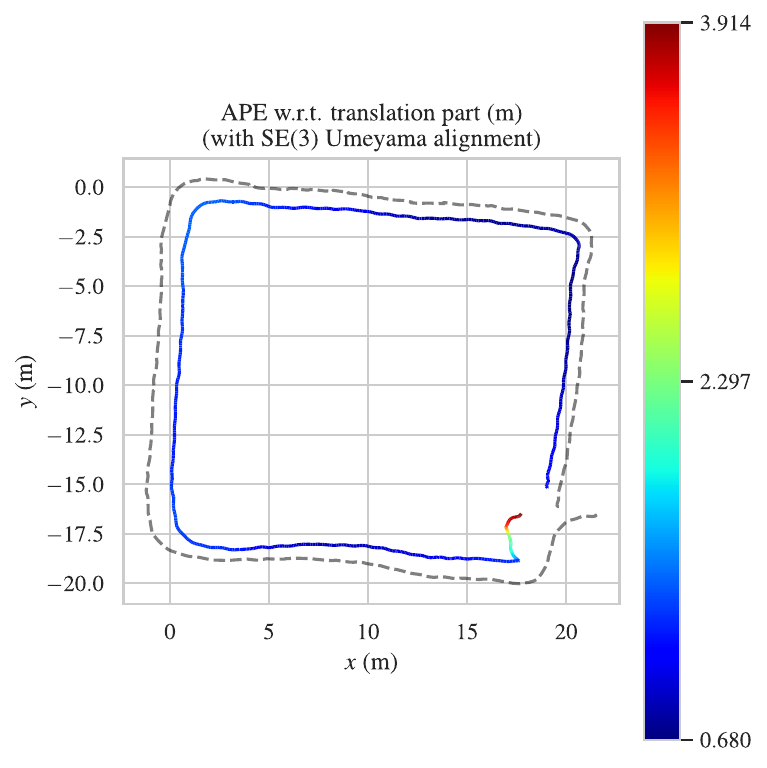}
        \caption{corridors-walk}
    \end{subfigure}
    \begin{subfigure}{0.3145\textwidth}
        \centering
        \includegraphics[height=0.3125\textheight]{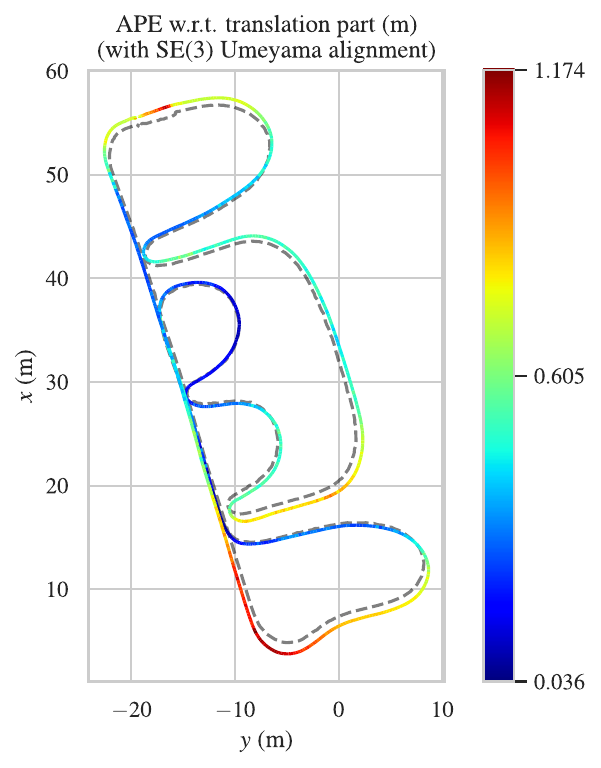}
        \caption{units-scooter}
    \end{subfigure}
    \begin{subfigure}{0.2705\textwidth}
        \centering
        \includegraphics[height=0.3125\textheight]{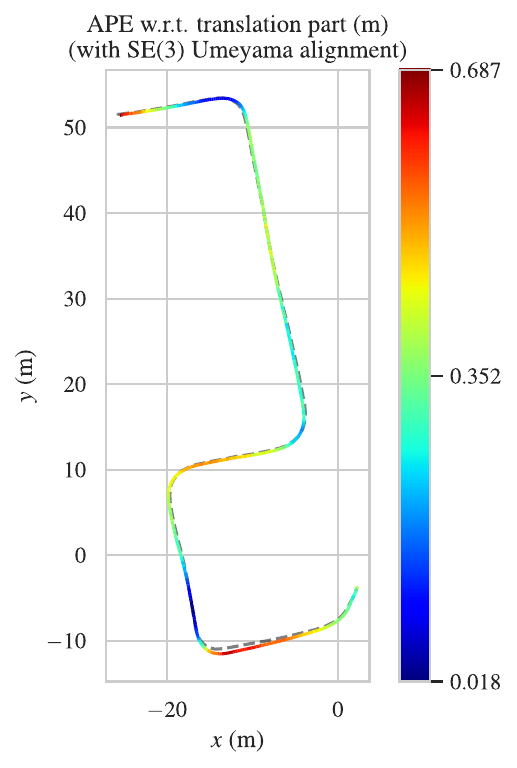}
        \caption{school-scooter}
    \end{subfigure}
    \caption{OKSVIS2 + SuperEvent's trajectories on the VECtor large-scale sequences with ground truth trajectory (dashed) and absolute position error of the 3D trajectories shown by the color.}
    \label{fig:vector}
    \vspace{25mm}
\end{figure*}

Figure~\ref{fig:vector} shows the trajectories from SuperEvent + OKVIS2 on the VECtor~\cite{Gao22} large-scale sequences. We report RPE scores in Table~\ref{tab:vector_rpe}.

\begin{table}[t]
\begin{center}
\caption{RPE scores in cm for consecutive samples at 20 Hz of OKVIS2~\cite{Leu22} + SuperEvent on the VECtor large-scale sequences.\label{tab:vector_rpe}}
\small
\begin{tabularx}{\linewidth}{
>{\centering}X
>{\centering}X
>{\centering}X
>{\centering}X
>{\centering}X
>{\centering\arraybackslash}X
}
\toprule
corr.- & corr.- & units- & units- & school- & school- \\
dolly & walk & dolly & scooter & dolly & scooter \\
\midrule
16.72 & 14.50 & 23.10 & 42.40 & 21.66 & 45.78 \\
\bottomrule
\end{tabularx}
\vspace{-12pt}
\end{center}
\end{table}

\end{document}